\documentclass{article}

\usepackage{arxiv}

\usepackage[utf8]{inputenc} 
\usepackage[T1]{fontenc}    
\usepackage{hyperref}       
\usepackage{url}            
\usepackage{booktabs}       
\usepackage{amsfonts}       
\usepackage{nicefrac}       
\usepackage{microtype}      
\usepackage{lipsum}
\usepackage{graphicx}
\usepackage{tabularx}
\usepackage{authblk}
\usepackage{amsmath} 
\usepackage{booktabs}
\usepackage{diagbox}
\usepackage{multirow}
\usepackage{makecell}
\usepackage{textcomp}
\usepackage{cite}
\graphicspath{ {./images/} }

\title{Multi-needle Localization for Pelvic Seed Implant Brachytherapy based on Tip-handle Detection and Matching\thanks{This work has been accepted for publication in IEEE Journal of Biomedical and Health Informatics. \textcopyright{} 2026 IEEE. The final authenticated version is available online at: \url{https://doi.org/10.1109/JBHI.2026.3694701}.}}

\author[1]{Zhuo Xiao}
\author[1]{Fugen Zhou}
\author[1]{Jingjing Wang}
\author[1]{Chongyu He}
\author[1]{Bo Liu}
\author[2]{Haitao Sun}
\author[2]{Zhe Ji}
\author[2]{Yuliang Jiang}
\author[2]{Junjie Wang}
\author[3]{Qiuwen Wu}

\affil[1]{Image Processing Center, Beihang University, Beijing 100191, China}
\affil[2]{Department of Radiation Oncology, Peking University Third Hospital, Beijing 100191, China}
\affil[3]{Department of Radiation Oncology, Duke University Medical Center, Durham, NC 27710, USA}
\affil[ ]{\textbf{\texttt{bo.liu@buaa.edu.cn}}}

\begin{document}
\maketitle
\begin{abstract}
Accurate multi-needle localization in intraoperative CT images is crucial for optimizing seed placement in pelvic seed implant brachytherapy. However, this task is challenging due to poor image contrast and needle adhesion. This paper presents a novel approach that reframes needle localization as a tip-handle detection and matching problem to overcome these difficulties. An anchor-free network, based on HRNet, is proposed to extract multi-scale features and accurately detect needle tips and handles by predicting their centers and orientations using decoupled branches for heatmap regression and polar angle prediction. To associate detected tips and handles into individual needles, a greedy matching and merging (GMM) method designed to solve the unbalanced assignment problem with constraints (UAP-C) is presented. The GMM method iteratively selects the most probable tip-handle pairs and merges them based on a distance metric to reconstruct 3D needle paths. Evaluated on a dataset of 100 patients, the proposed method demonstrates superior performance, achieving higher precision and F1 score compared to segmentation-based baselines utilizing the nnUNet model, thereby offering a more robust and accurate solution for needle localization in complex clinical scenarios. The source code is available at: https://github.com/BUAAXZzz/CTNeedleLocalization.
\end{abstract}

\begin{center}
\textbf{Published in IEEE Journal of Biomedical and Health Informatics}\\
\textbf{DOI: 10.1109/JBHI.2026.3694701}
\end{center}


\section{Introduction}
\label{sec:introduction}
Colorectal cancer remains a significant global health challenge, ranking 3rd in incidence and 2nd in mortality worldwide\cite{brayGlobalCancerStatistics2024}. For locally recurrent pelvic disease, Seed Implant Brachytherapy (SIBT) has emerged as a highly effective treatment modality, owing to its capacity for highly localized dose delivery
and superior radiobiological effects\cite{wangCTguidedRadioactive125I2023,martinez-monge125iodineBrachytherapyColorectal1998}. The therapeutic efficacy of SIBT is fundamentally dependent on the quality of the dose distribution: the goal is to maximize the radiation dose to the target volume while strictly minimizing exposure to organs at risk (OARs). Consequently, the precise geometric arrangement of the implanted seeds is the critical determinant of treatment success.

In standard clinical workflows, needle paths and seed positions are determined preoperatively; however, the actual intraoperative geometry inevitably differs from the initial plan\cite{jiangSideEffectsCTguided2018,poloReviewIntraoperativeImaging2010a,nagIntraoperativePlanningEvaluation2001}. During insertion, needle positions frequently shift due to tissue resistance, patient movement, and organ deformation\cite{tongCTguided125IInterstitialBrachytherapy2017,stoneProstateGlandMotion2002}. Because radioactive seeds are characterized by steep dose gradients, even minor reconstruction errors can lead to substantial variations in the delivered dose to the target and surrounding OARs\cite{DEUFEL2025772}. Such discrepancies pose a severe clinical risk, potentially creating "cold spots" that compromise tumor control or "hot spots" that increase toxicity\cite{MILES2008206,jainIntraoperative3DGuidance2012}. Therefore, relying solely on preoperative plans is insufficient; accurate localization of the implanted needles using post-insertion intraoperative CT is paramount for validating and correcting the actual dose distribution.

CT-based needle localization serves as the foundation for intraoperative quality assurance, enabling clinicians to project planned sources onto actual needle paths and recompute the 3D dose distribution. If dosimetric deviations are identified, immediate adjustments can be made by modifying seed positions or inserting supplementary needles to ensure optimal coverage\cite{wangEfficacyDosimetryAnalysis2020}. Despite the critical importance of this step, needle localization in current clinical practice is predominantly performed manually. This manual process is not only labor-intensive and time-consuming but also highly dependent on operator experience, creating a significant bottleneck that hinders rapid decision-making in the operating room. Consequently, there is an urgent need for robust, automatic needle localization methods to streamline the workflow and ensure consistent treatment accuracy.

\begin{figure}[!t]
\centerline{\includegraphics[width=\columnwidth]{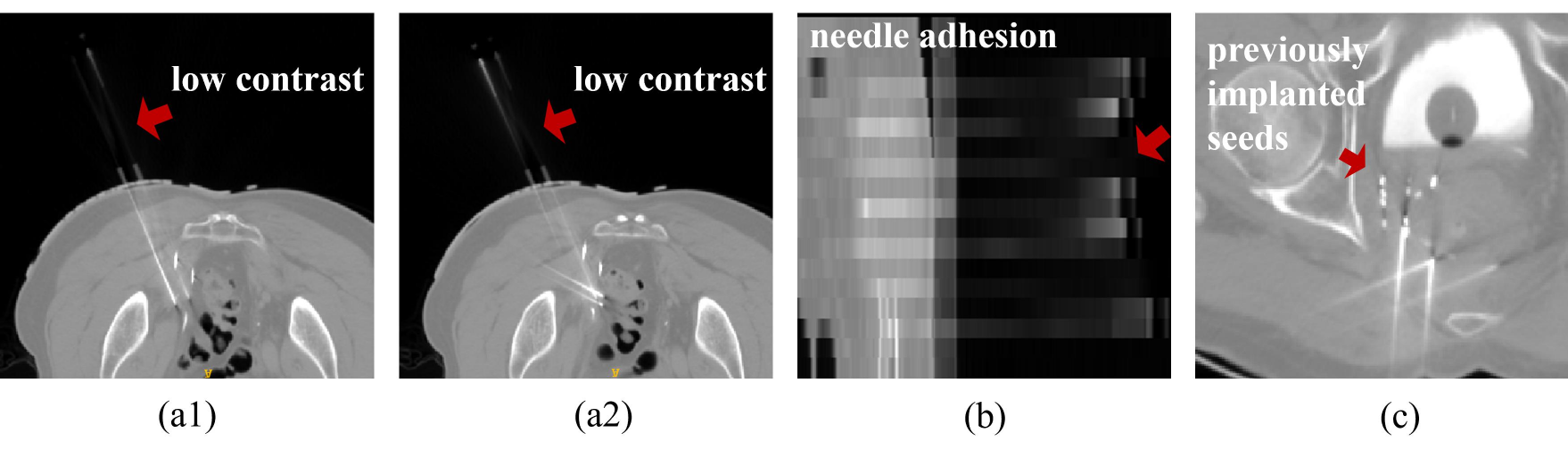}}
\caption{Challenges of multi-needle localization in intraoperative pelvic CT for seed implant brachytherapy. (a1)–(a2) Two consecutive axial slices where a needle shaft segment shows extremely low contrast (red arrows) due to partial-volume effects. (b) A sagittal slice illustrating needle overlap/adhesion between adjacent trajectories. (c) An axial slice from a recurrent rectal cancer case, where previously implanted seeds introduce strong streak artifacts that interfere with needle visualization.}
\label{fig1}
\end{figure}

However, reliable extraction of needle paths from intraoperative CT images remains technically challenging. One major difficulty arises from the partial-volume effect, which leads to uneven and obscure grayscale distributions along needle trajectories, complicating both manual delineation and automatic segmentation, as shown in Fig.~\ref{fig1}(a1) and (a2). Additionally, needle adhesion makes it difficult to separate individual needles (Fig.\ref{fig1}(b)). Furthermore, surrounding bone structures and previously-implanted seeds in cases of recurrent rectal cancer can have a significant impact on needle path localization (Fig.\ref{fig1}(c)).

While many studies have explored automatic needle localization, a significant gap remains in addressing the combined challenges of poor grayscale contrast and needle adhesion in complex multi-needle scenarios. Existing literature often focuses on perineal cranial to caudal needle trajectories, which are common in prostate and gynecological HDR-brachytherapy or biopsy procedures\cite{mehrtashAutomaticNeedleSegmentation2019, jungDeeplearningAssistedAutomatic2019, zhangMultiNeedleDetection3D2020, zhangAutomaticMultineedleLocalization2020, daiAutomaticMulticatheterDetection2020, andersenDeepLearningbasedDigitization2020, aleongRapidMulticatheterSegmentation2024, shaaerDeeplearningassistedAlgorithmCatheter2022, hrinivichSimultaneousAutomaticSegmentation2017}. In contrast, transperitoneal needle trajectories pose greater challenges for automatic localization. Moreover, many previous studies concentrate on single-needle detection, which is inherently less challenging than accurate localization in dense multi-needle settings\cite{mehrtashAutomaticNeedleSegmentation2019, zhouAutomaticNeedleSegmentation2008, zhouDeepLearningbasedAutomatic2024, hattEnhancedNeedleLocalization2015, barvaParallelIntegralProjection2008}. For lung seed implantation, Zheng et al. proposed a method combining thresholding and an improved RANSAC algorithm to extract and identify needle paths\cite{zhengAutomaticNeedleDetection2021}. However, this approach is confined to 2D slices with significant limitations. 

\begin{figure}[!t]
\centerline{\includegraphics[width=\columnwidth]{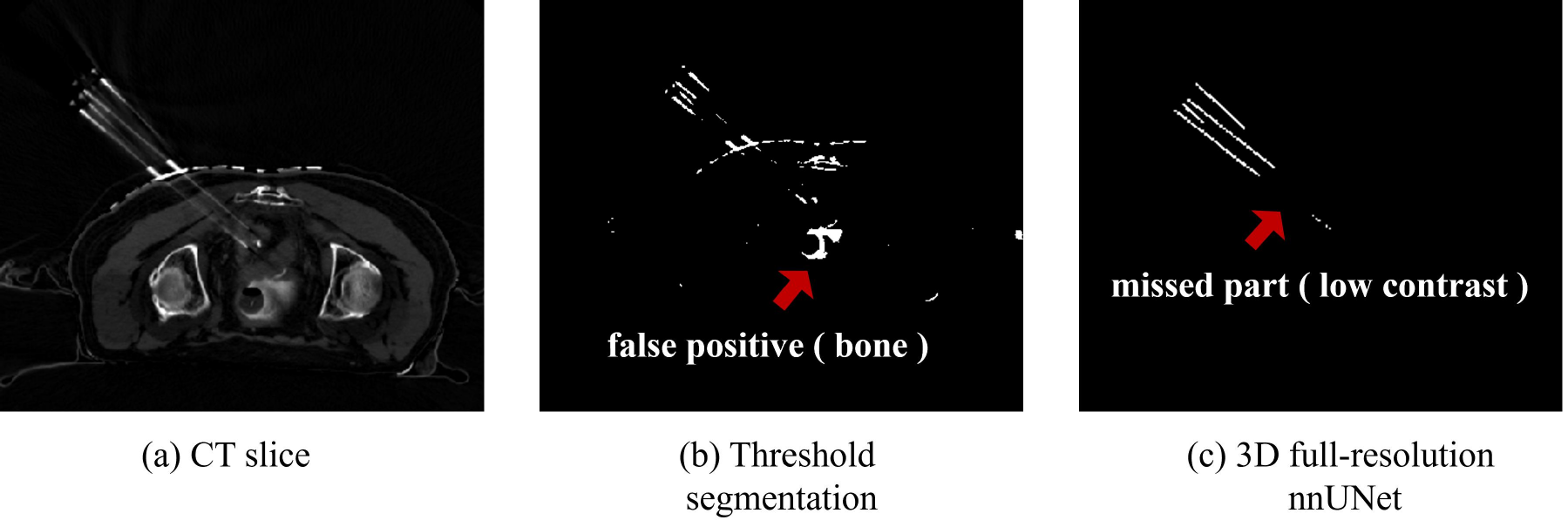}}
\caption{Demonstration of the insufficiency in segmentation results from both conventional thresholding and deep learning techniques. (a) One original axial CT slice. (b) Segmentation results using Otsu threshold\cite{otsuThresholdSelectionMethod1979}. (c) Segmentation results using deep learning-based method\cite{isenseeNnUNetSelfconfiguringMethod2021}.}
\label{fig2}
\end{figure}

Among existing methods, segmentation-based approaches are the most promising for multi-needle localization in pelvic SIBT. These methods typically involve segmenting needles using traditional thresholding or deep learning, followed by post-processing to identify needle trajectory instances\cite{zhengAutomaticNeedleDetection2021}. However, applying the segmentation-based method to the intraoperative brachytherapy CT remains challenging, as shown in Fig.\ref{fig2}, which illustrates the insufficient segmentation results obtained with both conventional thresholding\cite{otsuThresholdSelectionMethod1979} and deep learning techniques\cite{ronnebergerUNetConvolutionalNetworks2015,isenseeNnUNetSelfconfiguringMethod2021}. Furthermore, even with optimal segmentation, accurately extracting individual needle paths still remains difficult using traditional post-processing methods, such as curve fitting or connected component analysis due to the adhesion between needle paths.

Given these challenges, a deeper understanding of needle trajectory characteristics in CT images is essential. We observed that the needle handle, with its high contrast and visibility due to minimal external interference, provides a reliable reference. Similarly, the abrupt grayscale variations caused by metal artifacts\cite{barrettArtifactsCTRecognition2004} at the needle tip provide a distinct feature that can be leveraged for precise needle path localization.

To address the multi-needle localization problem, we propose a two-step approach: First, a tip-handle detection network is developed to detect needle tips and handles on axial slices. This network comprises a backbone for high-level semantic feature extraction and three anchor-free detection heads for tip-handle heatmap, local offset, and needle angle prediction. Second, the detected 2D tips and handles are matched to determine the 3D positions of needle tips and handles, and subsequently, the linear 3D needle paths. The matching problem is formulated as an unbalanced assignment problem with constraints (UAP-C), which we solve using a greedy matching and merging (GMM) method. This method is, to our knowledge, the first of its kind applied to multi-needle localization.

In summary, the contributions of this paper are as follows:

\begin{itemize}
    \item We pioneer a new paradigm for multi-needle localization by reframing it as a tip-handle detection and matching problem, which involves detecting needle tips and handles and pairing them accurately.
    
    \item We develop an anchor-free detection network, which predicts the centers and orientations of tips and handles through decoupled branches for heatmap regression and polar angle prediction.
    
    \item We introduce a greedy matching and merging method to solve the unbalanced assignment problem with constraints, iteratively selecting the most probable tip-handle pairs and merging them based on a distance metric to reconstruct individual needles.
    
    \item We annotated tips, handles, and needle paths in an intraoperative CT dataset of 100 patients, and experimental results demonstrate that our approach outperforms segmentation-based methods in precision and recall while achieving comparable accuracy in localization and orientation.
\end{itemize}

\section{Related Works}
\subsection{Object Detection}
Object detection aims to identify and classify objects within an image. In recent years, deep learning-based methods have dominated this field, primarily categorized into CNN-based and transformer-based approaches\cite{zouObjectDetection202023}. CNN-based object detection can be further divided into two-stage\cite{renFasterRCNNRealTime2017} and one-stage\cite{redmonYouOnlyLook2016,law2018cornernet, duanCenterNetKeypointTriplets2019, liuSSDSingleShot2016} methods. Two-stage detectors first generate region proposals\cite{renFasterRCNNRealTime2017} and then classify and refine their positions, ensuring high accuracy. One-stage detectors, on the other hand, bypass the proposal step and directly predict object locations and categories in a single pass, making them faster but sometimes less precise. DETR (Detection transformer)\cite{carionEndtoEndObjectDetection2020} revolutionizes object detection by treating it as a set prediction problem, which employs a transformer encoder-decoder architecture to directly predict object classes and bounding boxes in one step. However, DETR models typically require a large amount of training data and tend to converge slowly, making them less suitable for scenarios with limited data or real-time applications. This is particularly relevant in medical imaging datasets, which are typically much smaller in scale compared to those used for natural scenes. Therefore, our method adopts a CNN-based approach, which is more efficient in data-limited settings. Furthermore, in our case, the needle tips and handles have relatively fixed sizes, meaning that detecting their center points and orientations is more relevant than bounding boxes. To achieve this, we propose an anchor-free network that incorporates an angle prediction branch to fit the characteristics of the task.

\subsection{Assignment Problem}
The assignment problem (AP) is a core combinatorial optimization challenge, seeking to optimize the allocation of tasks to agents on a one-to-one basis, typically by minimizing total cost or maximizing total benefit. For standard, small-scale AP instances, established algorithms such as the Hungarian algorithm\cite{kuhnHungarianMethodAssignment1955} and linear programming\cite{ziliaskopoulosLinearProgrammingModel2000} provide optimal solutions. However, the generalized assignment problem (GAP)\cite{maniezzoMatheuristicsAlgorithmsImplementations2021} introduces resource constraints, such as agent capacities\cite{mazzolaGeneralizedAssignmentNonlinear1989}, making the problem NP-hard. Consequently, heuristic\cite{sahuSolvingAssignmentProblem, galadikovaUsingSimulatedAnnealing2024} and metaheuristic\cite{kinastHybridMetaheuristicSolution2022} approaches are frequently employed for practical, large-scale applications. More recently, machine learning\cite{schaferCombiningMachineLearning2023} and reinforcement learning\cite{hamzehiCombinatorialReinforcementLearning2019} have emerged as promising tools for tackling complex assignment problems, offering adaptive and efficient solutions in domains such as the Weapon-Target Assignment (WTA) problem\cite{klineWeaponTargetAssignmentProblem2019} and the Multi-Resource Generalized Assignment (MRGA) problem\cite{zhuDeepReinforcementLearning2018}. Nevertheless, these learning-based methods often lack rigorous theoretical performance guarantees and necessitate substantial training time, posing limitations in scenarios requiring rapid deployment or real-time responsiveness.

\subsection{Imaging and Preprocessing Techniques Relevant to Needle Localization}
In parallel with needle-localization methods, several imaging and preprocessing techniques also contribute to needle path extraction. Image registration frameworks\cite{swamidasImageRegistrationContour2020,czajkowskiAccuracyRegistrationsConebeam2020,rodgers3DTransrectalUltrasound2017} have been developed to align preoperative CT or MRI with intraoperative CT, cone-beam CT, or ultrasound in interstitial brachytherapy. Such co-registration enables the planned needle trajectories to be mapped onto the intraoperative anatomy, thereby offering an initial geometric prior and enabling quantitative assessment of insertion deviations for adaptive guidance. However, registration is never perfectly accurate; non-rigid deformations and residual alignment errors introduce unavoidable geometric discrepancies. Furthermore, registration cannot account for the independent shifting and bending of needles that occurs during insertion. Given that severe metal artifacts often make visual verification impossible, registration alone fails to provide the precision required for reliable multi-needle reconstruction.

From a complementary perspective, advanced CT metal artifact reduction algorithms have been investigated as a potential means to improve needle visibility in CT images. Both physics-driven\cite{gjestebyMetalArtifactReduction2016} and deep learning approaches\cite{wangInDuDoNetDeepUnfolding2023,karageorgosDenoisingDiffusionProbabilistic2024}, aim to suppress streak artifacts induced by metallic instruments and to recover more reliable Hounsfield Unit (HU) patterns around needles, which can improve the performance of needle localization. However, aggressive artifact reduction can inadvertently blur high-frequency details or alter the Hounsfield Units of the needle tips themselves, complicating their detection. Therefore, we utilize standard CT reconstructions without MAR. Our approach circumvents the artifact problem by focusing on tips and handles—which remain distinguishable even in noisy scans—and leveraging global geometric constraints to reconstruct the needle paths. 

\section{Methods}
\subsection{Overview}

\begin{figure*}[htbp]
    \centering
    \includegraphics[width=0.95\textwidth]{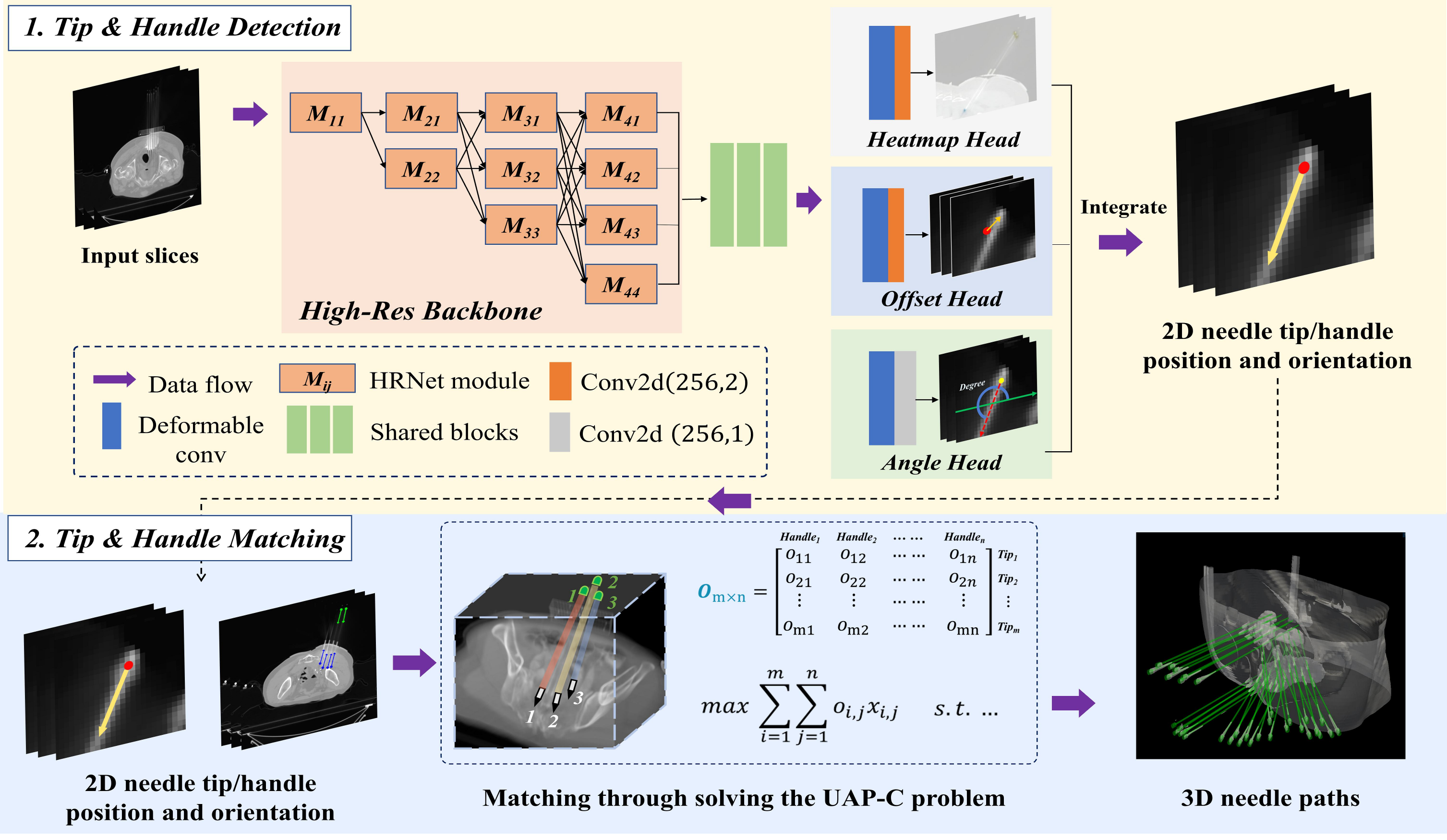} 
    \caption{The overall framework of the proposed method} 
    \label{fig:fig3} 
\end{figure*}

Fig. \ref{fig:fig3} illustrates the overall framework of the proposed method which consists of a tip-handle detection module and a tip-handle matching module. The detection module is designed to detect the needle tips and handles in 2D axial slices. It extracts multi-scale features through a high-resolution encoder, and predicts the position of the needle tips and handles on the 2D axial slices through three prediction heads, including the heatmap head, the offset head, and the angle head. Based on the detection results, the tip-handle matching module matches the detected tips and handles to determine the 3D needle path. The matching problem is formulated as an unbalanced assignment problem with constraints and is solved using a greedy matching and merging method.

For the tip-handle detection, the proposed detection framework uses a high-resolution backbone to extract multi-scale features, which are concatenated, processed through shared blocks, and then fed into three prediction heads (the heatmap head, the offset head, and the angle head). The heatmap head predicts a 2-channel heatmap that represents the probability of a pixel being the needle tip and handle. Local maxima within the 8-neighborhood of each pixel in the heatmap are then searched and those local maxima whose values exceed a predefined threshold are considered center point candidates for tips and handles. The offset head predicts an offset vector $(\Delta x, \Delta y)$ to compensate for quantization errors between the input image and the down-sampled heatmap resolution, ensuring more precise localization of the center points. The angle head predicts a polar angle $\theta$ for each pixel, which represents the 2D orientation of the needle related to the detected tip and handle in axial slices. This angle is defined as the direction of the vector connecting the tip or handle to the opposite end, measured relative to the X-axis, within the range $[0, 2\pi)$. Detection is performed in a 2D manner across all CT slices, and by integrating the results from three heads, the 3D positions $(x_o, y_o, z_o)$ and orientation $\alpha_o$ of the detected tips and handles are determined.

Based on the predicted information of needle tip and handle, all possible pairs of needle tips and handles are established. By incorporating a priori constraints and an objective function, a UAP-C is formulated. The proposed tip-handle matching strategy is then applied to merge duplicates and pair each needle tip with its corresponding handle, ensuring precise localization of 3D needle instances. The details of matching strategy will be described in subsequent sections.

\subsection{Detection of Needle Tips and Handles}
\subsubsection{Network Architecture }
The tip-handle detection network comprises a feature extractor and three output heads. The feature extractor uses HRNet-W48\cite{sunDeepHighResolutionRepresentation2019} as the backbone, which consists of four stages with four parallel resolution branches (48, 96, 192, and 384 channels from high to low resolution) and preserves high-resolution representations throughout the network, making it particularly effective for detecting small target objects such as needle tips and handles. After extracting multi-scale features through the backbone network, low-resolution feature maps are up-sampled to align with high-resolution feature maps and concatenated together, which are then fed into three convolutional blocks. The convolutional blocks are shared for the following three output heads, designed to minimize network redundancy and complexity while maintaining detection accuracy. Each convolutional block consists of a convolutional layer, a batch normalization layer, and a ReLU activation function, with the convolutional layer using a $3 \times 3$ kernel. The HRNet backbone itself follows the original implementation without structural modifications; all task-specific design is confined to the three prediction heads described below.

The output features of the shared convolutional blocks are passed to three distinct output heads, each responsible for different tasks: a heatmap prediction head for handles and tips, a local offset head, and an angle prediction head. Since the sizes of the needle handles and tips are fixed and not the primary focus of this study, the size branch is dropped, which typically predicts object dimensions in traditional detection frameworks\cite{zhou2019objects,law2018cornernet,nguyenCircleRepresentationMedical2022}.

Each head consists of a deformable convolution~\cite{dai2017deformable} and a standard $3 \times 3$ convolution. Unlike regular convolutions that operate on a fixed spatial grid, deformable convolutions dynamically adjust their sampling points, enabling adaptive and flexible feature extraction. Specifically, in a regular $3 \times 3$ convolution, the 9 sampling locations are fixed. In deformable convolutions, the positions of these sampling points are shifted by learned offsets $\Delta \mathbf{p} = (\Delta x, \Delta y)$, allowing the convolution to adapt to the object's geometry and capture more relevant features.

\subsubsection{Loss Function}
To train our detection network, we define a multi-task loss function consisting of three items:
\begin{equation}
L_{\text{det}} = \lambda_{\text{HM}} L_{\text{HM}} + \lambda_{\text{off}} L_{\text{off}} + \lambda_{\text{ang}} L_{\text{ang}}
\label{eq:det_loss}
\end{equation}
where $L_{\mathrm{HM}}$ is the heatmap loss, $L_{\mathrm{off}}$ is the offset loss, and $L_{\mathrm{ang}}$ is the angle prediction loss. $\lambda_{\mathrm{HM}}$, $\lambda_{\mathrm{off}}$, and $\lambda_{\mathrm{ang}}$ are the weighting factors for each loss term, respectively.

The ground truth of heatmap for needle tip and handle is calculated based on a 2D Gaussian kernel centered at the manually picked center points of the tips or handles. Focal loss~\cite{linFocalLossDense2020} is adopted to optimize the pixel-level logistic regression of heatmap:
\begin{align}
L_{\text{HM}} &= \frac{-1}{N} \sum_{(x,y,c)} 
\begin{cases} 
P_{(x,y)}^c & \text{if } H_{(x,y)}^c = 1 \\
Q_{(x,y)}^c & \text{otherwise}
\end{cases} \label{eq:heatmap_loss}
\end{align}
where $P_{(x,y)}^c = (1 - \hat{H}_{(x,y)}^c)^\alpha \log(\hat{H}_{(x,y)}^c)$, $Q_{(x,y)}^c = (1 - H_{(x,y)}^c)^\beta (\hat{H}_{(x,y)}^c)^\alpha \log(1 - \hat{H}_{(x,y)}^c)$, $\hat{H}_{(x,y)}^c$ denotes the predicted value at pixel $(x, y)$ for class $c$, $H_{(x,y)}^c$ is the ground truth value, $\alpha$ and $\beta$ are weighting hyper-parameters, and $N$ is the number of key points.

The offset layer is intended to correct the position inaccuracies after downsampling. The center location $(x, y)$ in the original image is mapped to the location $\left(\left\lfloor \frac{x}{d} \right\rfloor, \left\lfloor \frac{y}{d} \right\rfloor\right)$ in the heatmaps, where $d$ is the down-sampling factor. The offset $o_k$ and offset loss $L_{\text{off}}$ are defined as follows:
\begin{equation}
o_k = \left( \frac{x_k}{d} - \left\lfloor \frac{x_k}{d} \right\rfloor, \frac{y_k}{d} - \left\lfloor \frac{y_k}{d} \right\rfloor \right)
\label{eq:offset}
\end{equation}
\begin{equation}
L_{\text{off}} = \frac{1}{N} \sum_{k=1}^N \text{Smooth}_{L1}(o_k, \hat{o}_k)
\label{eq:offset_loss}
\end{equation}
where smooth $L1$ loss is given by:
\begin{equation}
\text{Smooth}_{L1}(x_i, y_i) = \begin{cases} 
0.5 (x_i - y_i)^2 & \text{if } |x_i - y_i| < 1 \\
|x_i - y_i| - 0.5 & \text{otherwise}
\end{cases}
\label{eq:smooth_l1}
\end{equation}

The angle loss is used to regress the 2D orientation of tips and handles, predicting the polar angle at each pixel location in the heatmap. Due to the periodic nature of angles\cite{yangArbitraryOrientedObjectDetection2020}, applying an $L1$ loss directly to the angle predictions is impractical, as it would introduce discontinuities at angular boundaries. Instead, a cosine loss is employed to address this periodicity:
\begin{equation}
L_{\text{ang}} = \frac{1}{N} \sum_{k=1}^N \left(1 - \cos(\hat{\theta}_k - \theta_k)\right)
\label{eq:angle_loss}
\end{equation}
Here, $\theta_k$ is the ground truth angle for the $k$-th object, and $\hat{\theta}_k$ is the predicted angle.
\subsection{Matching of Needle Tips and Handles}
After utilizing detection networks to identify the 2D positions and orientations of needle handles and tips across all slices, a tip-handle matching algorithm is proposed to match the corresponding handles and tips to determine the 3D needle paths. We formulate the problem as an unbalanced assignment problem with constraints and propose to solve it through a greedy matching and merging method.

\subsubsection{The Unbalanced Assignment Problem with Constraints}
The assignment problem is a fundamental combinatorial optimization problem that involves assigning an equal number of tasks to agents, with the goal of minimizing total cost or maximizing total benefit. The UAP-C extends the assignment problem, where the number of tasks and agents may not be equal, and the number of possible matchings may be less than the number of tasks and agents. Furthermore, additional constraints are imposed during the assignment process.
In the context of our problem, the tasks represent needle tips, while the agents are needle handles. Table \ref{tab:tip_handle_matching} presents the notation used in the UAP-C model for the tip-handle matching problem.

\begin{table}[!t]
\centering
\caption{The description of sets, parameters, and variables used in the tip-handle matching problem.}
\label{tab:tip_handle_matching}
\begin{tabular}{>{\raggedright\arraybackslash}p{3cm} p{5cm}}
\toprule
\textbf{Function / Symbol} & \textbf{Description} \\
\midrule
\texttt{NoCross} & No two needles in the solution set intersect, which can be checked using the shortest distance between two line segments. \\
\midrule
\textbf{Sets} & \\
\midrule
$T=\{1,\ldots,m\}$ & Index set of detected needle tips \\
$H=\{1,\ldots,n\}$ & Index set of detected needle handles \\
\midrule
\textbf{Parameters} & \\
\midrule
$o_{i,j}$ & Score value for pairing tip $i$ and handle $j$ \\
$L_{i,j}$ & 3D Euclidean distance between the tip $i$ and handle $j$ (mm) \\
$L_{\text{prior}}$ & Prior needle length (mm) \\
$\varepsilon_l$ & Tolerance for needle length difference (mm) \\
$\text{at}_i$ & Detected angle for tip $i$ ($^\circ$) \\
$\text{ah}_j$ & Detected angle for handle $j$ ($^\circ$) \\
$\varepsilon_a$ & Tolerance for angle difference ($^\circ$) \\
$N_{\text{prior}}$ & A priori number of implanted needles \\
\midrule
\textbf{Variable} & \\
\midrule
$x_{i,j}$ & 1 if tip $i$ is matched with handle $j$, and 0 otherwise \\
\bottomrule
\end{tabular}
\end{table}

Using these definitions, the tip-handle matching problem can be mathematically formulated as follows:

\begin{align}
\max \quad & \sum_{i \in T} \sum_{j \in H} o_{i,j}\, x_{i,j} \label{eq:objective} \\
\text{s.t.}\quad
& \sum_{i \in T} x_{i,j} \le 1, \forall j \in H; \quad \sum_{j \in H} x_{i,j} \le 1, \forall i \in T \label{eq:constraints8} \\
& \bigl|L_{i,j} - L_{\text{prior}}\bigr| < \varepsilon_l \qquad \forall i \in T, j \in H \label{eq:constraint10} \\
& \bigl|\mathrm{at}_i - \mathrm{ah}_j\bigr| < \varepsilon_a \qquad \forall i \in T, j \in H \label{eq:constraint11} \\
& \mathrm{NoCross}(i,j,k,l) \qquad \forall (i,j)\neq(k,l) \label{eq:constraint12} \\
& x_{i,j} \in \{0,1\} \qquad \forall i \in T, j \in H \label{eq:constraint13} \\
& \sum_{i \in T} \sum_{j \in H} x_{i,j} \le N_{\text{prior}} \label{eq:constraint14}
\end{align}

In this model, the objective is to maximize the total score of all matched tip-handle pairs. Considering that the HU along a needle shaft is not only relatively high but also spatially consistent, we define the score function by analyzing the HU statistics along each candidate tip–handle line segment, so that bright yet strongly fluctuating paths (e.g., induced by streak artifacts or previously implanted seeds) are naturally down-weighted. Let $t_i$ and $h_j$ denote the center positions of the $i$-th detected needle tip and the $j$-th detected needle handle, respectively. The score $o_{i,j}$ for pairing tip $t_i$ and handle $h_j$ is defined as:
\begin{equation}
o_{i,j} = \frac{\mu_{i,j}}{\sigma_{i,j}}
\label{eq:score_function}
\end{equation}
where $\mu_{i,j}$ and $\sigma_{i,j}$ denote the average and standard deviation of HU on the line segment, respectively.

Constraint \ref{eq:constraints8} enforces a unique matching, ensuring that each handle is assigned to at most one tip and vice versa. Constraint \ref{eq:constraint10} ensures that the length of each tip-handle pair closely approximates the prior needle length. Constraint \ref{eq:constraint11} excludes pairs with inconsistent angles at the tip and handle, while Constraint \ref{eq:constraint12} prevents any two needles from intersecting. Finally, Constraint \ref{eq:constraint13} defines the decision variables as binary, and Constraint \ref{eq:constraint14} limits the total number of selected needle paths to the prior needle count.

\subsubsection{The Greedy Matching and Merging Method}
Several methods have been developed to solve the assignment problems. The Hungarian algorithm, designed for balanced problems, can be adapted for unbalanced ones by adding dummies, but struggles with complex constraints. Mathematical programming methods like integer and linear programming relaxation has advantage to find the optimality. However, in real-world complex problems, they face issues such as exponential computational complexity and difficulty in formulating complex constraints. Heuristic algorithms like genetic and greedy algorithms are useful alternatives. Genetic algorithms handle complex constraints well in large-scale problems, while greedy algorithms make fast, locally-optimal choices, providing practical solutions for real-world scenarios.

For the above tip-handle matching UAP-C with complex constraints, we propose solving it using a greedy matching and merging strategy. Initially, a score matrix $S$, similar to that used in the Hungarian method, is constructed to facilitate the pairing of tip-handle candidates. Each element of the score matrix $S_{ij}$ is the score value of the potential needle path connecting the $i$-th tip and the $j$-th handle. To enforce constraints \ref{eq:constraint10} and \ref{eq:constraint11}, candidate pairs that do not satisfy the prescribed criteria are assigned a score of negative infinity ($-\infty$). In this way, geometrically invalid pairings are excluded from subsequent selection, so that only feasible candidates are retained for matching.

An iterative greedy matching procedure is then applied to match feasible tip-handle pairs. At each iteration, the needle path with the highest finite score that does not intersect with any previously selected paths is chosen. This approach ensures that the non-intersection constraint is respected at every step. In the case of an $m \times n$ score matrix, applying the greedy matching process results in up to $\min(m,n)$ matching pairs between the $m$ needle tips and $n$ needle handles. Let $x$ denote the number of selected matching pairs. Given the known total number of implanted needles, $N_{\text{prior}}$, the current matching solution is accepted when $x \leq N_{\text{prior}}$. Conversely, if $x > N_{\text{prior}}$, this indicates the presence of duplicate needle paths and the following merging process is performed.

The merging process involves combining both the tip and handle of duplicate needle paths into a new, unified position. To identify duplicates, the algorithm first targets needle paths with lower score values. A low score typically suggests insufficient grayscale intensity along the path or poor intensity uniformity, which is often caused by duplicate paths that reduce the overall score. For needle paths with lower scores, if the tips and handles of two paths are within $2.5\,\mathrm{mm}$ of each other, they are classified as duplicates, and their respective tips and handles are merged. The merging process uses the masks of the tips and handles detected by our detection network. Specifically, for the tip (though the same procedure applies to the handle), when two duplicate tips are identified, the average non-zero HU values within their respective detected mask regions are calculated. This average HU value is used as a weight to determine the new tip position, which is computed as the weighted sum of the two previous positions. This ensures that the new tip location reflects the weighted contribution of both original positions. The same procedure is applied to merge the handles of duplicate paths. The merging process is performed iteratively, refining the solution until the number of remaining needle paths is equal to $N_{\text{prior}}$, the known target number of implanted needles.

\section{Experiments and results}
\label{sec:exp}
\subsection{Patient Data and Preprocessing}
\label{sec:preprocess}
One hundred intraoperative CT images were collected from colorectal cancer patients who had undergone SIBT. The study was approved by the Institutional Ethics Committee of Peking University Third Hospital (Beijing, China; Approval No. M2021438). The cohort included 66 males and 34 females, with a median age of 56 years (interquartile range, 49–64 years; range, 27–84 years). According to the AJCC 8th edition staging system\cite{weiserAJCC8thEdition2018}, there were 9 stage I, 31 stage II, 53 stage III, and 7 stage IV cases. All 100 patients underwent SIBT for locally recurrent pelvic disease after prior surgery and/or chemoradiotherapy; no primary pelvic disease cases were included in this retrospective cohort.

All needle tips and handles within each CT volume were manually annotated by a clinical expert and verified by a senior radiation oncologist. These annotations served as the ground truth for both model training and evaluation. Specifically, due to their finite size and the partial volume effect, the needle tip or handle may span multiple slices. Since a 2D network is used for tip-handle detection, all imaging areas of the handle or tip were annotated as the ground truth for training the detection network using ITK-SNAP software, in order to avoid misleading the network during learning. To validate the final matching results, the annotated 2D handles and tips were fused to obtain 3D positions, thereby determining the 3D needle paths. The 3D positions of the tip and handle were calculated as the intensity-weighted centroids of the annotated 2D handles and tips.

All cases used standardized brachytherapy needles (18G $\times$ 200\,mm); we set $L_{\text{prior}} = 205\,\text{mm}$ (tip-to-handle center distance) and applied a tolerance of $\varepsilon_l = 10\,\text{mm}$. $N_{\text{prior}}$ was obtained from the surgical record. The tolerance window enhances robustness to minor deviations in the effective tip--handle distance arising from imaging resolution or measurement uncertainty. As long as the discrepancy remains within $\varepsilon_l$, correct matches are preserved, whereas substantial prior mis-specification beyond this range may lead to reduced recall. The number of needles per image ranged from 2 to 43, with an average of $15.2 \pm 8.2$. The CT images varied in resolution, with in-slice resolution ranging from $0.68\,\mathrm{mm}$ to $1.05\,\mathrm{mm}$ and slice thickness set at $5.0\,\mathrm{mm}$. The number of CT slices per image ranged from 7 to 29, with an average of $15.2 \pm 5.4$. For the development of the detection method, all experiments were conducted under the patient-wise five-fold cross-validation described in Section \ref{sec:implementation}.

An image preprocessing step is applied to the CT scan to enhance the visibility of the needle paths, which facilitates the subsequent tip-handle matching process. This is achieved through the white top-hat transformation, a technique that highlights smaller, brighter structures, such as the needle paths, while suppressing larger background elements. The white top-hat transformation was implemented using the SimpleITK toolkit\cite{lowekampDesignSimpleITK2013}, applied slice-by-slice. A spherical structuring element with a radius of 5 pixels was used, which is sufficiently large relative to the size of the needle path, ensuring that the transformation effectively isolates the needle paths from surrounding tissues. 

The performance of the proposed method was evaluated by comparison with two segmentation-based baselines, both of which relied on 3D full-resolution nnUNet model~\cite{isenseeNnUNetSelfconfiguringMethod2021} for initial needle segmentation. We selected segmentation-based pipelines because they represent the predominant and most promising paradigm for multi-needle localization in CT-guided brachytherapy as discussed in the Introduction. Subsequently, we determined the 3D tip and handle locations by calculating the intensity-weighted centroids of manually annotated 2D handle and tip regions across the slices they spanned within the segmentation masks. The ground truth for the needle path mask was established through iterative pseudo-label generation and refinement, culminating in manual correction. Initially, we created pseudo-labels by generating a 3-mm-radius 3D cylindrical mask between the annotated needle tip and handle. These served to train the 3D-fullres nnUNet model using its default settings, which produced more precise labels. We repeated this process twice, applying manual corrections in the final iteration to obtain the definitive ground truth.

\subsection{Implementation Details}
\label{sec:implementation}
\subsubsection{Network Training}
The tip-handle detection network was developed using PyTorch and trained on an NVIDIA RTX 3090 GPU. The model was optimized using Adam with an initial learning rate of $2.5 \times 10^{-4}$, a batch size of 8, and a step decay schedule for 100 epochs. Data augmentation, including cropping, scaling, shifting, and horizontal/vertical flipping, was applied randomly to improve the stability of the training and the generalization ability of the model. The weight parameters for the loss functions $\lambda_{HM}$, $\lambda_{off}$ and $\lambda_{ang}$ are set to 2, 1 and 1, respectively. 

For heatmap generation and heatmap loss, we empirically set the standard deviation of the Gaussian kernel $\sigma=\text{radius}/3$, $\alpha = 2$ and $\beta = 4$ in our experiments following the setting of previous works \cite{zhou2019objects,law2018cornernet}. All CT slices were processed at a resolution of $512 \times 512$, and the radii were defined in mm and converted to pixels according to the in-plane spacing of each volume. Specifically, the radii of the needle tip and handle were set to 3 mm and 4 mm, respectively, at the input image scale for heatmap generation. These values were empirically selected to approximate the effective CT footprint of the needle components, reflecting their physical size and imaging effects such as metal blooming artifacts. This alignment of radii with object geometry was designed to promote optimization stability according to prior works \cite{law2018cornernet}, while a sensitivity analysis with ±1 mm perturbations also demonstrated the robustness of the proposed framework to this parameter, with relative changes in F1 within 1.5\% and localization and angle MAE changes within 5\%, indicating that the detector is robust to reasonable radius variations.

To enhance the contrast of the needle tip and handle, CT intensities were thresholded at 800 HU, with all values above 800 HU clipped to this upper bound, and then linearly normalized to the interval $[-1, 1]$.

\subsubsection{Evaluation metrics}
To evaluate the proposed approach, a five-fold cross-validation was conducted. The 100 patients were randomly divided into five groups, each containing twenty patients. In each validation fold, four groups of patients were used to train the detection network, and the remaining group was used to validate tip-handle detection and matching. The evaluation framework includes both the 2D detection of needle tips and handles and the subsequent 3D localization of needles.

To quantitatively evaluate the 2D detection network for needle tips and handles, five metrics were employed: mean absolute error ($\mathrm{MAE}_{\mathrm{pos2d}}$ in mm), angle error for tips and handles ($\mathrm{MAE}_{\mathrm{agl2d}}$ in $^{\circ}$), recall, precision, and F1 score. Manual annotations served as the ground truth. A detection was considered correct if the center distance between the predicted and ground truth positions was within $2\,\mathrm{mm}$ and the angle difference was within $5^\circ$.

The localization error, $MAE_{pos2d}$, was calculated as the average absolute distance between the 2D predicted position ($\hat{p}_i$) and the 2D ground truth position ($p_i$) across all tips or handles:
\begin{equation}
MAE_{pos2d} = \frac{1}{n} \sum_{i = 1}^{n} \left| \hat{p}_i - p_i \right|
\end{equation}

The angular error for the needle tip or handle was calculated by averaging the absolute differences between the predicted ($\hat{\theta}_i$) and the ground truth angle ($\theta_i$):

\begin{equation}
MAE_{agl2d} = \frac{1}{n} \sum_{i = 1}^{n} \left| \hat{\theta}_i - \theta_i \right|
\end{equation}

Here, $n$ denotes the number of detected tips or handles.

Similarly, five metrics were employed to evaluate the performance of 3D needle localization. Firstly, standard detection evaluation metrics including precision, recall, and F1 score were utilized to assess the performance of the proposed method. A successful needle detection was defined as follows: the Euclidean distances between both the detected needle tip and handle positions and their corresponding ground truth positions must be within $2.5\,\mathrm{mm}$, and this detected needle must represent the closest match to the ground truth. The $2.5\,\mathrm{mm}$ threshold was set according to the slice thickness ($5\,\mathrm{mm}$) of the dataset.

Secondly, for the correctly detected needles, the localization accuracies for the tip-handle position and needle orientation were evaluated. The location accuracy of the tip and handle position using the MAE of the 3D position ($MAE_{tip3d}$ and $MAE_{hdl3d}$) and the relative error in each axis ($MAE_{tip-relx}$, $MAE_{tip-rely}$, $MAE_{tip-relz}$, $MAE_{hdl-relx}$, $MAE_{hdl-rely}$, $MAE_{hdl-relz}$) to give a more thorough assessment of the location accuracy. Let $\hat{P}_i = \{\hat{P}_{i,x},\hat{P}_{i,y},\hat{P}_{i,z}\} \in \mathbb{R}^3$ represent the ground truth position of the needle tip or handle for the $i$-th instance, $P_i = \{P_{i,x},P_{i,y},P_{i,z}\} \in \mathbb{R}^3$ represent the detected position of the needle tip for the same instance and $n$ be the number of matched needles. With $\{s_x, s_y, s_z\}$ representing the spacing of the CT, the $MAE_{tip3d}$ and $MAE_{tip-relx}$ were calculated as:
\begin{equation}
MAE_{tip3d} = \frac{1}{n} \sum_{i = 1}^{n} \left| \hat{P}_i - P_i \right|
\label{eq:mae_tip3d}
\end{equation}

\begin{equation}
MAE_{tip\text{-}relx} = \frac{1}{n} \sum_{i = 1}^{n} \frac{\left| \hat{P}_{i,x} - P_{i,x} \right|}{s_x}
\label{eq:mae_tip_relx}
\end{equation}

The other MAEs were calculated similarly.

The localization accuracy of the needle orientation was evaluated via the error of the needle path angle, calculated as:
\begin{equation}
MAE_{agl3d} = \frac{1}{n} \sum_{i = 1}^{n} \arccos\left( \hat{A}_i \cdot A_i \right)
\label{eq:mae_agl3d}
\end{equation}

where $A_i$ and $\hat{A}_i$ denote the 3D orientation of the ground truth and detected needle.
\subsection{Evaluation of Needle Tip and Handle Detection}
Table \ref{tab:detection_results} presents the quantitative results of detection network. We further compared the proposed framework with recent state-of-the-art learning-based detectors. For the 2D detection task, these include horizontal box-based detectors, specifically RT-DETR\cite{zhaoDETRsBeatYOLOs2024a} and DeiM\cite{huang2025deim}, as well as the oriented detector PKINet\cite{caiPolyKernelInception2024a}. All learning-based detectors were trained and evaluated under the same data split and preprocessing as the proposed method. For horizontal detectors, the center of the predicted bounding box was used as the endpoint location for computing $MAE_{pos2d}$. Orientation error was only reported for methods that explicitly predict orientation (PKINet and the proposed method), and was computed on true-positive detections.

As shown in Table \ref{tab:detection_results}, the proposed detector achieves the best overall F1 scores for both needle tips and handles, while maintaining sub-1.5 mm 2D positional errors. Compared with recent learning-based detectors (RT-DETR, DeiM, and PKINet), our method is specifically tailored to endpoint localization by predicting keypoint centers and orientations rather than relying on bounding-box regression. For box-based baselines (RT-DETR and DeiM), the endpoint location is approximated by the bounding-box center, which is less consistent for slender, high-aspect-ratio objects and can be sensitive to box extent variations across adjacent slices. In contrast, the proposed heatmap–offset formulation directly supervises the endpoint center and therefore provides more stable localization under partial-volume effects and streak artifacts.

A consistent observation across methods is that handles exhibit larger $MAE_{pos2d}$ than tips, while their angular errors are smaller. This can be explained by the distinct imaging characteristics of the two endpoints in intraoperative pelvic CT. The handle region is usually more extended and may span a larger visible area outside the patient, making its “center” less well-defined and thus more susceptible to center-of-mass ambiguity, which amplifies positional deviations. Meanwhile, the handle often presents a clearer elongated structure with higher contrast and less severe metal-induced distortion than the tip region, enabling more reliable orientation cues and leading to lower $MAE_{agl2d}$. Conversely, needle tips are frequently affected by strong metal artifacts and local contrast reduction near bones or previously implanted seeds, which can blur the endpoint appearance and increase the uncertainty of the orientation estimation.

Finally, the recall values for both tips and handles are consistently high, indicating that the network rarely misses true endpoints under the adopted TP criteria. Precision is slightly lower than recall, which is mainly attributed to false positives caused by high-density structures (e.g., bone boundaries and seed-induced artifacts) that can mimic endpoint-like patterns on individual slices. Notably, these slice-level false positives can be further suppressed in the subsequent 3D tip–handle matching stage by enforcing geometric consistency (length/angle constraints) and excluding implausible pairings, which contributes to the improved robustness of the full localization pipeline.

\begin{table*}[!t]
\centering
\caption{Detection results of needle handles and needle tips.}
\label{tab:detection_results}
\normalsize
\begin{tabular}{llccccc}
\toprule
\textbf{Method} & \textbf{Object} &
\textbf{MAE$_{\text{pos2d}}$ (mm)} &
\textbf{MAE$_{\text{agl2d}}$ ($^\circ$)} &
\textbf{Recall} &
\textbf{Precision} &
\textbf{F1} \\
\midrule
\multirow{2}{*}{RT-DETR~\cite{zhaoDETRsBeatYOLOs2024a}}
& Tip    & $1.26 \pm 0.24$ & -- & $92.9 \pm 2.8$ & $88.3 \pm 7.8$ & $89.6 \pm 3.9$ \\
& Handle & $1.49 \pm 0.56$ & -- & $97.9 \pm 1.1$ & $86.5 \pm 6.9$ & $91.5 \pm 4.2$ \\
\midrule
\multirow{2}{*}{DeiM}~\cite{huang2025deim}
& Tip    & $1.28 \pm 0.28$ & -- & $93.4 \pm 2.7$ & $85.7 \pm 5.3$ & $88.3 \pm 3.7$ \\
& Handle & $1.46 \pm 0.45$ & -- & $98.4 \pm 1.7$ & $85.0 \pm 5.4$ & $90.6 \pm 3.6$ \\
\midrule
\multirow{2}{*}{PKINet~\cite{caiPolyKernelInception2024a}}
& Tip    & $1.22 \pm 0.23$ & $3.85 \pm 1.36$ & $90.5 \pm 1.5$ & $88.2 \pm 6.2$ & $88.7 \pm 4.3$ \\
& Handle & $1.45 \pm 0.46$ & $3.67 \pm 0.50$ & $93.3 \pm 1.5$ & $86.9 \pm 7.0$ & $89.6 \pm 2.4$ \\
\midrule
\multirow{2}{*}{Proposed}
& Tip    & $1.25 \pm 0.33$ & $3.75 \pm 2.10$ & $99.4 \pm 1.7$ & $91.4 \pm 8.3$ & $95.0 \pm 4.7$ \\
& Handle & $1.46 \pm 0.61$ & $2.12 \pm 0.76$ & $99.0 \pm 2.0$ & $93.1 \pm 7.7$ & $95.8 \pm 4.6$ \\
\bottomrule
\end{tabular}
\end{table*}

\begin{figure}[htbp]
    \centering
    \includegraphics[width=\columnwidth]{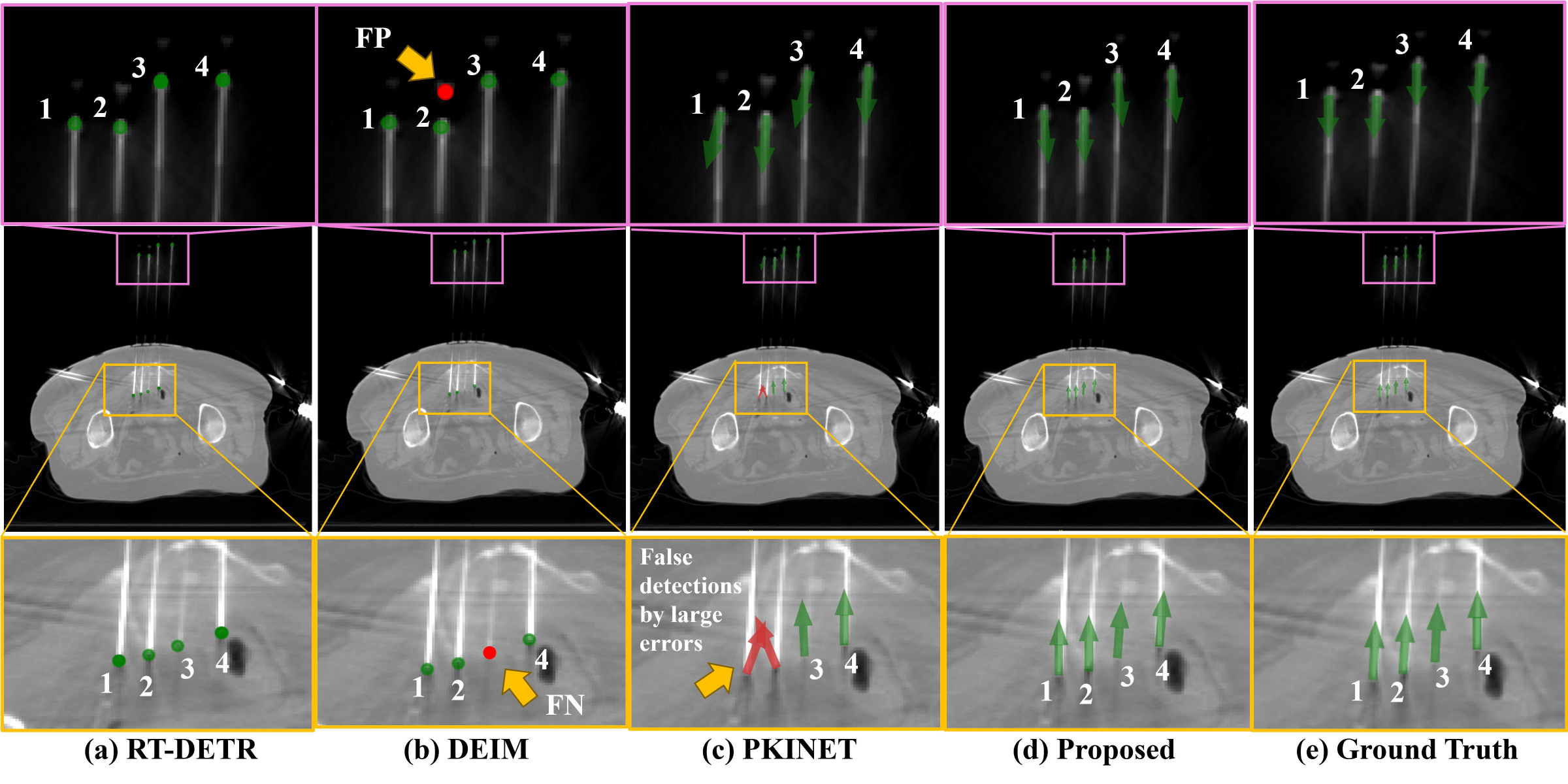}
    \caption{Qualitative comparison of 2D detection results across different methods. Needle handle and tip regions marked by the gray and yellow boxes are zoomed in and then displayed above and below to show details of detection. In every subfigure, true positives are marked in green; false positives and false negatives are marked in red.}
    \label{fig:fig4}
\end{figure}

From Fig. \ref{fig:fig4}, we qualitatively compare 2D endpoint detection across RT-DETR, DeiM, PKINet, and the proposed method. RT-DETR and DeiM output horizontal bounding boxes, while the clinically consideration is the center location of the needle tip or handle. Therefore, we visualize their predictions as center dots. In contrast, both PKINet and the proposed method explicitly predict endpoint orientation, and their results are therefore visualized using arrows to indicate both position and direction. Overall, the DeiM tends to miss subtle needle tips and occasionally produce false positives near the superior part of the needle handle. PKINet often localizes the correct center, but its orientation can be unstable. In these cases, large angle deviations cause otherwise well-localized detections to be evaluated as false positives under the joint position-and-orientation criterion.

\subsection{Evaluation of 3D Needle Localization}
Table~\ref{tab:needle_localization} summarizes the 3D needle localization performance of the proposed method compared to two segmentation-based baselines: S\&M, which denotes a Segmentation \& Manual pipeline where needles are segmented by nnU-Net and individual trajectories are manually extracted, and S\&R, which denotes a Segmentation \& RANSAC pipeline that applies iterative RANSAC line fitting\cite{zhengAutomaticNeedleDetection2021,zhangAutomaticMultineedleLocalization2020}) to nnU-Net segmentation outputs\cite{isenseeNnUNetSelfconfiguringMethod2021}.

The proposed method demonstrated superior overall performance, particularly in terms of Precision and F1 score. Specifically, compared to the segmentation-based baseline, our method achieved a significantly higher precision ($92.7 \pm 9.6\%$ vs.\ $87.1 \pm 8.7\%$, $p < 0.01$) and F1 score, while maintaining a comparable recall. In contrast, the RANSAC baseline exhibited substantially degraded performance across all detection metrics (Recall, Precision, and F1) compared to the proposed method ($p < 0.01$), together with a noticeably larger variance. This instability mainly arises in cases with template-guided insertions and large target volumes, where multiple needles are nearly parallel and densely spaced. In such scenarios, segmented voxels from neighboring needles are often spatially interleaved, leading the S\&R method to fit incorrect trajectories.

Regarding geometric accuracy for correctly localized needles, the proposed framework significantly improved tip localization performance. As shown in Table \ref{tab:needle_localization}, the proposed method achieved a lower 3D tip error ($\mathrm{MAE}_{\mathrm{tip3d}}$) of $1.05 \pm 0.36$~mm compared to $1.34 \pm 0.51$~mm for the segmentation-based baseline ($p = 0.02$). Decomposed analysis reveals that this improvement is primarily driven by reduced errors along the x-axis ($p = 0.07$) and z-axis ($p = 0.04$). Meanwhile, the handle localization ($\mathrm{MAE}_{\mathrm{hdl3d}}$) and path angle estimation ($\mathrm{MAE}_{\mathrm{agl3d}}$) errors, showed no statistically significant differences between the proposed method and the segmentation baseline (all $p > 0.1$). These results suggest that the primary advantage of the proposed tip-handle detection framework lies in its ability to suppress false positives and enhance tip localization robustness in the presence of pelvic intraoperative CT interference.

\begin{table*}[htbp]
\centering
\caption{3D needle localization results.}
\label{tab:needle_localization}
\normalsize

\begin{tabular}{lcccccc}
\toprule
\textbf{Method} &
\textbf{Recall} &
\textbf{Precision} &
\textbf{F1} &
$\textbf{MAE}_{\textbf{tip3d}}$ &
$\textbf{MAE}_{\textbf{tip-relx}}$ &
$\textbf{MAE}_{\textbf{tip-rely}}$ \\
\midrule
Segmentation\&Manual
& $86.5 \pm 9.1$ & $87.1 \pm 8.7$ & $86.8 \pm 8.7$
& $1.34 \pm 0.51$ & $0.84 \pm 0.41$ & $0.78 \pm 0.32$ \\
Segmentation\&RANSAC~\cite{zhengAutomaticNeedleDetection2021,zhangAutomaticMultineedleLocalization2020})
& $79.3 \pm 20.2$ & $78.6 \pm 20.1$ & $78.9 \pm 20.1$
& $1.74 \pm 0.62$ & $0.70 \pm 0.38$ & $1.13 \pm 0.63$ \\
Proposed
& $88.6 \pm 11.5$ & $92.7 \pm 9.6$ & $90.3 \pm 9.6$
& $1.05 \pm 0.36$ & $0.64 \pm 0.27$ & $0.66 \pm 0.20$ \\
\midrule
$p$ value (S\&M vs P)
& 0.13 & $<\!0.01$ & 0.03 & 0.02 & 0.07 & 0.35 \\
$p$ value (S\&R vs P)
& $<\!0.01$ & $<\!0.01$ & $<\!0.01$ & $<\!0.01$ & 0.20 & $<\!0.01$ \\
\bottomrule
\end{tabular}

\vspace{1em}

\begin{tabular}{lcccccc}
\toprule
\textbf{Method} &
$\textbf{MAE}_{\textbf{tip-relz}}$ &
$\textbf{MAE}_{\textbf{hdl3d}}$ &
$\textbf{MAE}_{\textbf{hdl-relx}}$ &
$\textbf{MAE}_{\textbf{hdl-rely}}$ &
$\textbf{MAE}_{\textbf{hdl-relz}}$ &
$\textbf{MAE}_{\textbf{agl3d}}$ \\
\midrule
Segmentation\&Manual
& $0.14 \pm 0.10$ & $1.61 \pm 0.88$ & $0.77 \pm 0.44$
& $1.18 \pm 0.80$ & $0.15 \pm 0.13$ & $0.30 \pm 0.08$ \\
Segmentation\&RANSAC~\cite{zhengAutomaticNeedleDetection2021,zhangAutomaticMultineedleLocalization2020})
& $0.16 \pm 0.09$ & $2.28 \pm 0.51$ & $0.69 \pm 0.47$
& $1.61 \pm 0.63$ & $0.22 \pm 0.09$ & $0.47 \pm 0.16$ \\
Proposed
& $0.10 \pm 0.08$ & $1.73 \pm 0.95$ & $0.75 \pm 0.42$
& $1.19 \pm 0.87$ & $0.16 \pm 0.16$ & $0.26 \pm 0.06$ \\
\midrule
$p$ value (S\&M vs P)
& 0.04 & 0.94 & 0.58 & 0.74 & 0.50 & 0.13 \\
$p$ value (S\&R vs P)
& 0.05 & $<\!0.01$ & 0.52 & 0.25 & $<\!0.01$ & $<\!0.01$ \\
\bottomrule
\end{tabular}

\vspace{0.5em}
\raggedright
\footnotesize
Note: S\&M: Segmentation \& Manual; S\&R: Segmentation \& RANSAC; P: Proposed Method. The $p$ value is calculated using a two-sided Wilcoxon signed-rank test.
\end{table*}

\noindent\textbf{Sensitivity to Needle Overlap/Adhesion:}
We further assessed robustness under dense configurations by stratifying needles into two subsets according to the ground-truth annotations: with overlap/adhesion vs.\ without overlap/adhesion. The proposed method maintained comparable detection and localization performance between subsets. Specifically, the F1 score remained stable ($87.8 \pm 8.9\%$ vs.\ $91.0 \pm 3.3\%$, $p=0.36$), with similar precision ($92.6 \pm 8.7\%$ vs.\ $93.7 \pm 3.0\%$, $p=0.88$) and recall ($84.4 \pm 9.4\%$ vs.\ $88.8 \pm 3.6\%$, $p=0.39$). Geometric localization accuracy was also comparable between the two groups. No statistically significant differences were observed in 3D tip localization error, endpoint position errors along individual axes, handle localization accuracy, or needle orientation error (two-sided Wilcoxon signed-rank test, all $p > 0.05$), indicating that the proposed tip–handle detection and matching formulation is robust to needle overlap and adhesion in challenging multi-needle scenarios.

\begin{figure}[htbp]
    \centering
    \includegraphics[width=\columnwidth]{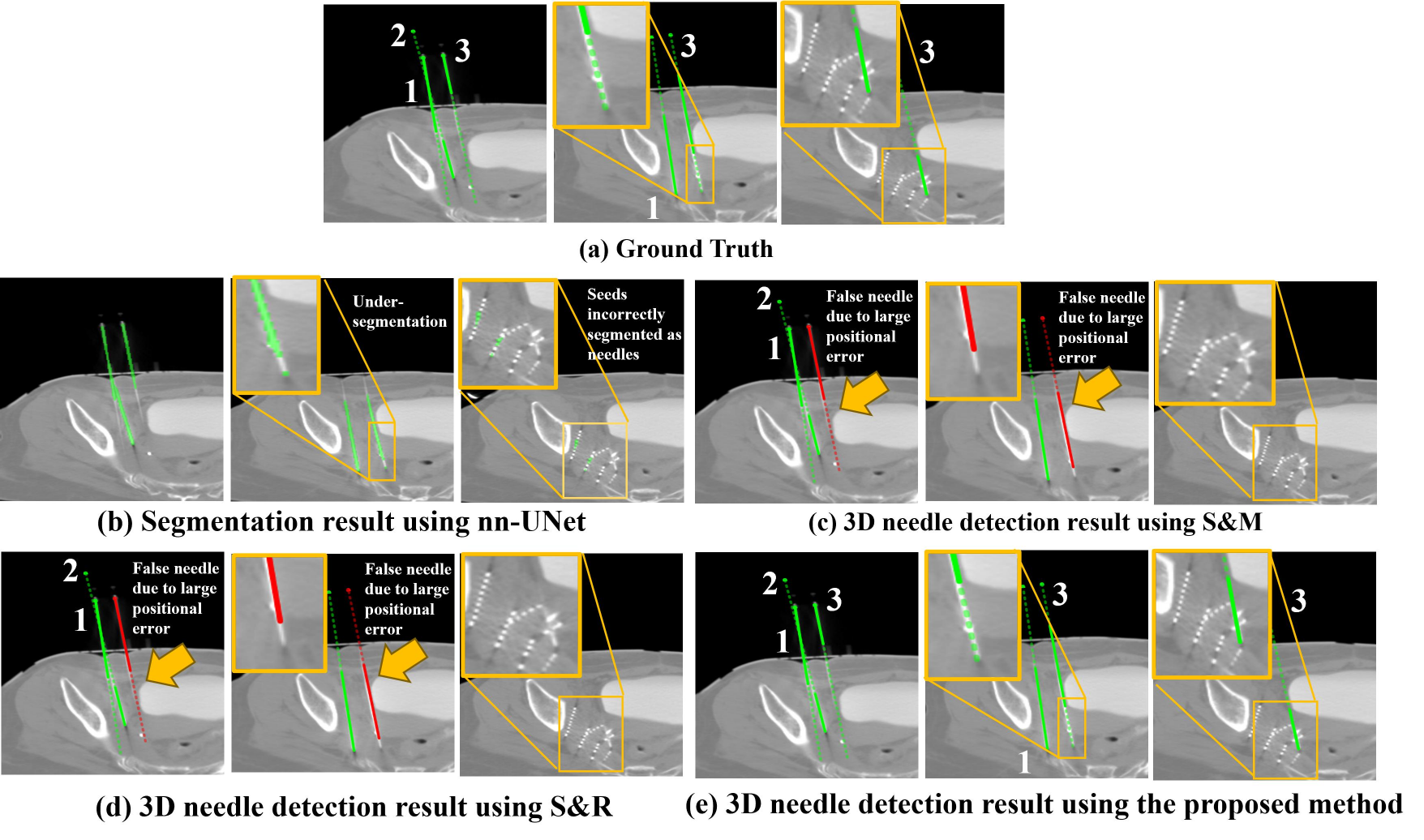} %
    \caption{Segmentation and detection results on three consecutive slices for a patient case. Correctly detected needles are shown in green, while red indicates a false positive needle. Yellow boxes indicate regions of interest and are shown in zoomed-in views for clarity. Solid lines indicate the portion of the needles intersecting the current image slice. Dotted lines represent parts of the correctly detected needles located on other slices.} 
    \label{fig:fig5} 
\end{figure}

Fig.\ref{fig:fig5} shows the segmentation and detection results on three consecutive slices for a patient case. The proposed method correctly localizes all needles, showing consistent trajectories across slices. However, the S\&M and the S\&R baseline incorrectly detected the needle indicated by the yellow arrow, which spans three slices. This failure, consistent with the criteria discussed earlier, stems from a missing tip segmentation on the final slice due to seed interference. In contrast, our proposed tip-handle matching strategy demonstrates greater robustness and successfully detects this needle.

\begin{figure}[!t]
\centerline{\includegraphics[width=\columnwidth]{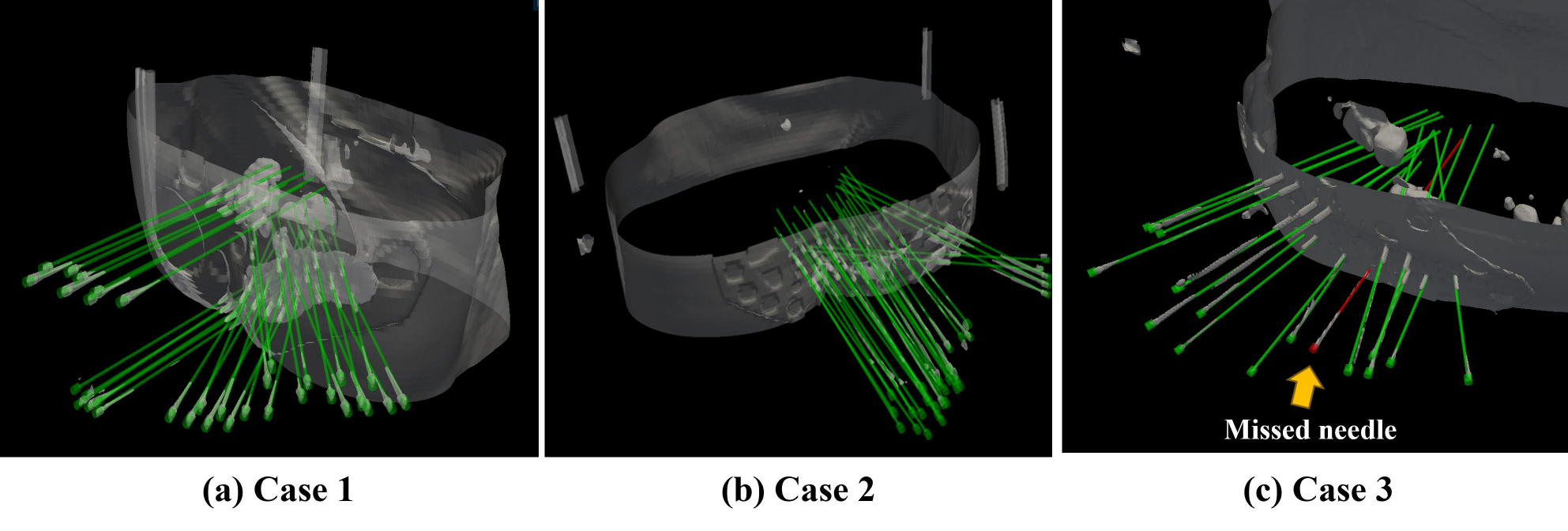}}
\caption{3D visualization of the detected needles of the proposed method along with the iso-surface (-400 HU) of the CT for three cases. Green lines denote correctly detected needles true positives, while red lines indicate missed needle trajectories false negatives, as highlighted by the yellow arrow.}
\label{fig6}
\end{figure}
Fig. \ref{fig6} presents the 3D visualization of the detection result of the proposed method for three cases. While the needle visualization is somewhat unclear due to their subtle grayvalue distribution, the detected needles align well with the implanted needles, particularly evident around the needle guidance holes.

\subsection{Computational Efficiency}
Processing time is a critical factor in clinical intraoperative planning. On a personal computer equipped with an Intel Xeon\textregistered\ Gold 5118 CPU and an NVIDIA RTX 3090 GPU, the total computing time was $9.41 \pm 8.95$ seconds per case. The inference time for detecting the needle tip and handle was much quicker, at $1.83 \pm 0.12$ seconds, largely depending on the number of CT slices of the case. The majority of the computation time was spent on the matching process, which took an average of $7.57 \pm 8.92$ seconds.

The computational complexity of the proposed GMM algorithm consists of candidate-pair generation and ranking. Let $m$ and $n$ denote the numbers of detected needle tips and handles, respectively. Constructing the score matrix $\mathbf{S} = [o_{i,j}] \in \mathbb{R}^{m \times n}$ requires $O(mn)$ time. After geometric filtering, let $K$ denote the number of retained tip--handle pairs. Ranking these pairs requires $O(K \log K)$ time, while the subsequent greedy matching step does not increase the overall asymptotic complexity. Thus, the overall complexity is $O(mn + K \log K)$, with a worst-case complexity of $O(mn \log (mn))$ when no candidate pair is removed by the pairwise geometric filtering.

In the tested cases, the most extensive matching process involved 44 needles and took 24 seconds, while cases with 5 matched needles were processed in 0.34 seconds. For comparison, in routine clinical practice, an experienced medical physicist typically requires approximately 4 minutes to manually delineate and associate needle trajectories slice by slice for a case with 16 needles. This contrast highlights that the proposed pipeline can substantially reduce the localization time while maintaining full automation, making it compatible with intraoperative workflows where rapid feedback is essential.

\subsection{Ablation Study}

\begin{table*}[!t]
\centering
\caption{Ablation study of the angle prediction branch for 2D needle detection.}
\label{tab:ablation_angle_2d}
\normalsize
\begin{tabular}{llccccc}
\toprule
\textbf{Method} & \textbf{Object} &
\textbf{MAE$_{\text{pos2d}}$ (mm)} &
\textbf{MAE$_{\text{agl2d}}$ ($^\circ$)} &
\textbf{Recall} &
\textbf{Precision} &
\textbf{F1} \\
\midrule
\multirow{2}{*}{w/o angle prediction}
& Tip    & $1.28 \pm 0.26$ & -- & $97.3 \pm 1.7$ & $90.2 \pm 8.0$ & $93.2 \pm 3.2$ \\
& Handle & $1.50 \pm 0.60$ & -- & $98.1 \pm 1.9$ & $93.3 \pm 8.0$ & $95.5 \pm 3.7$ \\
\midrule
\multirow{2}{*}{with angle prediction}
& Tip    & $1.25 \pm 0.33$ & $3.75 \pm 2.10$ & $99.4 \pm 1.7$ & $91.4 \pm 8.3$ & $95.0 \pm 4.7$ \\
& Handle & $1.46 \pm 0.61$ & $2.12 \pm 0.76$ & $99.0 \pm 2.0$ & $93.1 \pm 7.7$ & $95.8 \pm 4.6$ \\
\bottomrule
\end{tabular}
\end{table*}

\begin{table*}[htbp]
\centering
\caption{Ablation study of key components in the proposed tip–handle matching pipeline for 3D needle localization.}
\label{tab:ablation}
\normalsize

\begin{tabular}{lcccccc}
\toprule
\textbf{Configuration} &
\textbf{Recall} &
\textbf{Precision} &
\textbf{F1} &
$\textbf{MAE}_{\textbf{tip3d}}$ &
$\textbf{MAE}_{\textbf{tip-relx}}$ &
$\textbf{MAE}_{\textbf{tip-rely}}$ \\
\midrule
w/o angle constraint
& $82.9 \pm 17.4$ & $86.8 \pm 14.8$ & $84.5 \pm 15.8$
& $1.17 \pm 0.42$ & $0.62 \pm 0.26$ & $0.70 \pm 0.38$ \\
w/o NoCross constraint
& $79.7 \pm 17.1$ & $85.9 \pm 16.0$ & $82.5 \pm 16.3$
& $1.19 \pm 0.37$ & $0.66 \pm 0.30$ & $0.78 \pm 0.32$ \\
w/o merging step
& $79.7 \pm 24.2$ & $80.3 \pm 23.9$ & $79.9 \pm 24.0$
& $1.22 \pm 0.38$ & $0.65 \pm 0.43$ & $0.82 \pm 0.43$ \\
w/o white top-hat preprocessing
& $82.5 \pm 13.7$ & $88.4 \pm 11.6$ & $85.0 \pm 12.1$
& $1.15 \pm 0.37$ & $0.62 \pm 0.28$ & $0.67 \pm 0.23$ \\
Proposed
& $88.6 \pm 11.5$ & $92.7 \pm 9.6$ & $90.3 \pm 9.6$
& $1.05 \pm 0.36$ & $0.64 \pm 0.27$ & $0.66 \pm 0.20$ \\
\bottomrule
\end{tabular}

\vspace{1em}

\begin{tabular}{lcccccc}
\toprule
\textbf{Configuration} &
$\textbf{MAE}_{\textbf{tip-relz}}$ &
$\textbf{MAE}_{\textbf{hdl3d}}$ &
$\textbf{MAE}_{\textbf{hdl-relx}}$ &
$\textbf{MAE}_{\textbf{hdl-rely}}$ &
$\textbf{MAE}_{\textbf{hdl-relz}}$ &
$\textbf{MAE}_{\textbf{agl3d}}$ \\
\midrule
w/o angle constraint
& $0.13 \pm 0.11$ & $1.75 \pm 0.32$ & $0.76 \pm 0.27$
& $1.07 \pm 0.78$ & $0.20 \pm 0.23$ & $0.28 \pm 0.06$ \\
w/o NoCross constraint
& $0.10 \pm 0.07$ & $1.74 \pm 0.62$ & $0.74 \pm 0.22$
& $1.19 \pm 0.59$ & $0.16 \pm 0.13$ & $0.40 \pm 0.17$ \\
w/o merging step
& $0.15 \pm 0.11$ & $1.79 \pm 1.10$ & $0.76 \pm 0.22$
& $1.16 \pm 0.73$ & $0.22 \pm 0.16$ & $0.58 \pm 0.16$ \\
w/o white top-hat preprocessing
& $0.12 \pm 0.13$ & $1.71 \pm 0.83$ & $0.74 \pm 0.42$
& $1.15 \pm 0.84$ & $0.17 \pm 0.13$ & $0.29 \pm 0.07$ \\
Proposed
& $0.10 \pm 0.08$ & $1.73 \pm 0.95$ & $0.75 \pm 0.42$
& $1.19 \pm 0.87$ & $0.16 \pm 0.16$ & $0.26 \pm 0.06$ \\
\bottomrule
\end{tabular}
\end{table*}

An ablation study was performed to systematically evaluate the contribution of each key design component in the proposed 3D needle localization pipeline, with the results summarized in Table \ref{tab:ablation}. All ablation variants followed the same preprocessing, network training protocol, and metric definitions in Sections \ref{sec:preprocess} and \ref{sec:implementation}. The w/o angle branch variant is evaluated using 2D tip and handle detection metrics, while the other variants are evaluated using 3D needle localization metrics.

Specifically, for 2D needle tip-handle detection, we evaluate a variant without angle branch (w/o angle branch), by removing the angle prediction head from the 2D detection network and using position-only endpoint detection. For 3D tip-handle matching, several variants are evaluated, including (a) w/o angle constraint: removing the angle-consistency constraint in \ref{eq:constraint11}. In this setting, the downstream matching does not rely on angular information. (b) w/o NoCross constraint: removing the non-intersection constraint in \ref{eq:constraint12}. (c) w/o merging step: disabling the duplicate-path merging procedure when the greedy matching returns more than $N_{prior}$ paths; in this case, the algorithm directly keeps the top-ranked paths without merging duplicates. (d) w/o white top-hat preprocessing: computing the path score in \ref{eq:objective} on the original CT slices without the white top-hat enhancement described in Section \ref{sec:preprocess}.

Removing the angle prediction branch consistently degraded the 2D endpoint detection performance, as shown in Table~\ref{tab:ablation_angle_2d}. Specifically, the F1 score for needle tips decreased from $95.0\%$ to $93.2\%$, accompanied by an increase in localization error ($\mathrm{MAE}_{\mathrm{pos2d}}$ from $1.25$ mm to $1.28$ mm). A similar trend was observed for handles, where the F1 score decreased from $95.8\%$ to $95.5\%$. These results indicate that explicit angle supervision provides complementary geometric cues beyond positional information alone, thereby improving overall detection robustness. This finding also suggests that angular information contributes not only at the matching stage, but already at the endpoint detection stage by improving discrimination between true endpoints and anatomically or artifact-induced confounders.

As shown in Table~\ref{tab:ablation}, removing either the NoCross constraint or the merging step led to the most pronounced performance degradation. The F1 score dropped from $90.3\%$ to $82.5\%$ and $79.9\%$, respectively, together with reductions in both recall and precision, while the 3D tip error ($\mathrm{MAE}_{\mathrm{tip3d}}$) increased from $1.05$~mm to $1.19$~mm and $1.22$~mm. In particular, disabling the merging step markedly reduced precision (from $92.7\%$ to $80.3\%$), indicating that duplicate trajectories produced by greedy matching can occupy the limited path set and suppress true needles. This observation shows that the merging step is particularly important in dense multi-needle cases, where multiple high-scoring candidate paths may correspond to the same physical needle. Both variants also exhibited larger angular errors ($\mathrm{MAE}_{\mathrm{agl3d}} = 0.40^\circ$ and $0.58^\circ$), reflecting increased mismatching in dense configurations.

Removing the angle constraint mainly increased mismatches between tips and handles with similar appearance scores but inconsistent orientations, causing a noticeable decrease in F1 (from $90.3\%$ to $84.5\%$) and a larger orientation error ($\mathrm{MAE}_{\mathrm{agl3d}}$ from $0.26^\circ$ to $0.28^\circ$). Compared with removing the angle prediction branch in the detector, this result further indicates that angular information plays two complementary roles in the proposed pipeline: it improves 2D endpoint localization during detection and also regularizes 3D tip-handle association during matching. Finally, white top-hat preprocessing provided a modest but consistent gain in overall matching robustness, as it stabilized the intensity statistics along candidate paths used by Eq.~\ref{eq:score_function}, especially for low-contrast needles. The corresponding improvement is smaller than that of the geometric constraints, suggesting that preprocessing mainly serves as a supportive component that improves score reliability under challenging image appearance, whereas the dominant performance gains come from the geometry-aware design of the detection and matching strategy.

\subsection{Failure Cases and Potential Remedies}
\begin{figure}[!t]
\centerline{\includegraphics[width=\columnwidth]{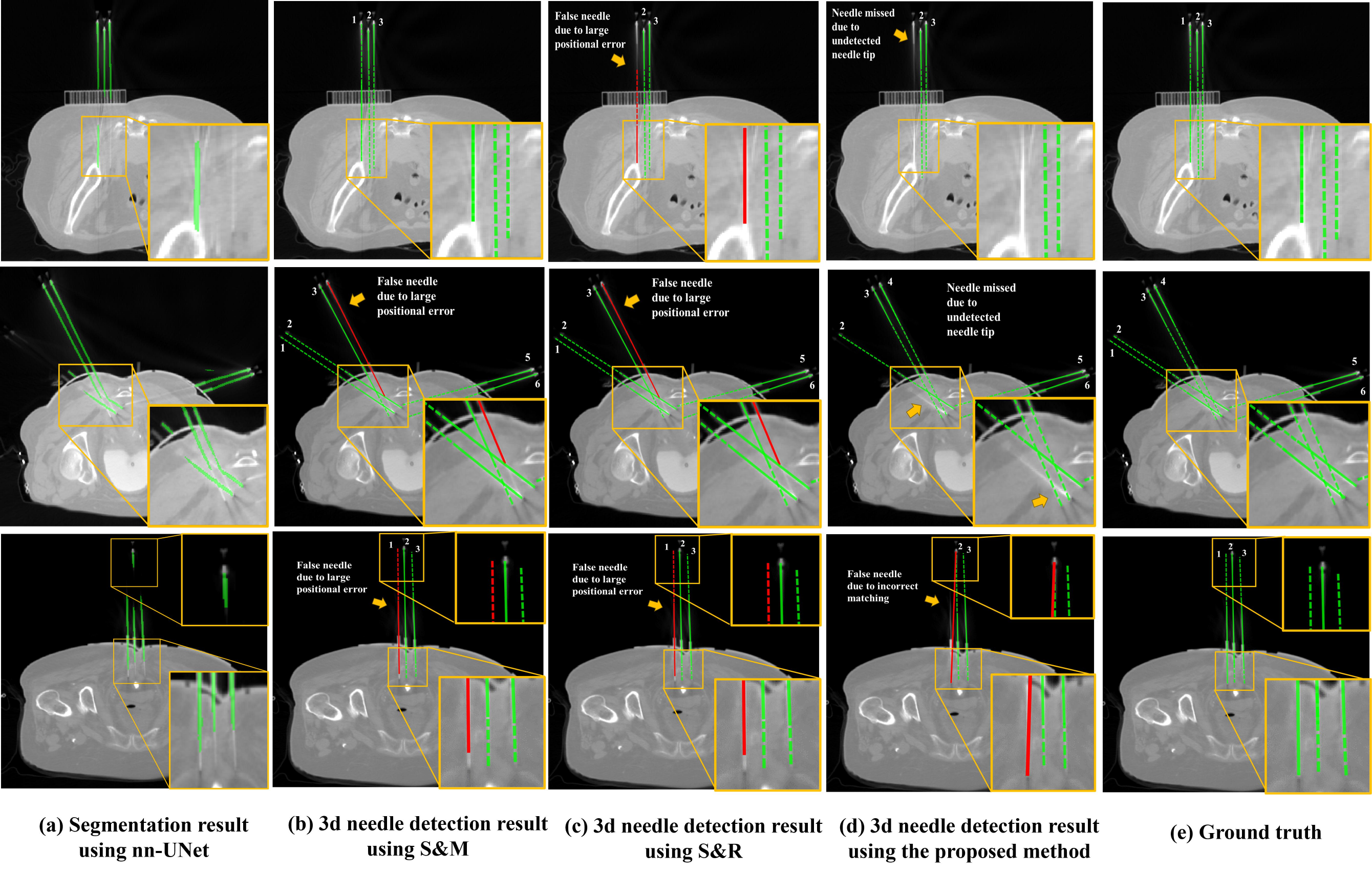}}
\caption{Representative failure cases of the proposed needle localization method for three patient cases. Each row represents a single patient case, and each column displays the results of a specific method. Correctly detected needles are shown in green, while red indicates a false positive needle. Solid lines indicate the portion of the needles intersecting the current image slice. Dotted lines represent parts of the correctly detected needles located on other slices. Yellow boxes indicate regions of interest and are shown in zoomed-in views for clarity.}
\label{fig7}
\end{figure}

Although the proposed pipeline is robust in most cases, a few challenging scenarios may still lead to failures. Fig.~\ref{fig7} summarizes three representative failure patterns and compares our method with segmentation-based baselines.

In the first case, the needle tip is located close to high-density bone, where streak artifacts and locally reduced contrast weaken the characteristic tip appearance. As a result, the proposed detector misses the tip endpoint, and the subsequent tip--handle association fails to reconstruct the corresponding trajectory (Fig.~\ref{fig7}, row~1(d)). For the segmentation result, nnU-Net may not produce a fully continuous shaft mask under such weak visibility conditions (Fig.~\ref{fig7}, row~1(a)). Nevertheless, when partial responses are still present near the tip and handle regions, a human observer can often infer plausible endpoints by exploiting the fragmented mask and approximate collinearity, leading to trajectories consistent with the ground truth (Fig.~\ref{fig7}, row~1(b)). In contrast, RANSAC-based line fitting fails when adjacent shafts are tightly clustered, as voxels from neighboring needles become spatially interleaved, which can induce incorrect fitted trajectories and result in both false positives and false negatives (Fig.~\ref{fig7}, row~1(c)).

A similar failure mode is observed in dense multi-needle layouts. In this case, the tip region of needle~\#4 becomes partially superimposed onto needle~\#1 due to partial-volume effects, which suppresses the distinctive tip cue of needle~\#1 and leads to a missed tip detection and a false negative in our pipeline (Fig.~\ref{fig7}, row~2(d)). For segmentation-based approaches, severe under-segmentation is observed around shaft crossing or adhesion regions, making reliable separation of individual trajectories difficult. This degradation propagates to downstream reconstruction, causing failures for both the S\&M and S\&R fitting strategies in crowded regions (Fig.~\ref{fig7}, row~2(a--c)).

In the third case, the failure corresponds to a false-positive trajectory caused by an incorrect tip--handle association (Fig.~\ref{fig7}, row~3(d)). Specifically, the handle of needle~\#1 is not detected, and its tip is mistakenly matched to a cross-slice handle candidate belonging to needle~\#2, resulting in a spurious reconstructed path. In contrast, the S\&M and S\&R baselines produce erroneous detections mainly due to under-segmentation, which subsequently leads to incorrect trajectory extraction (Fig.~\ref{fig7}, row~3(a--c)).

From a treatment-planning perspective, false negatives may lead to missing needles, while false positives may introduce spurious ones; both scenarios can compromise intraoperative dose calculations. In addition, errors in the position or orientation of reconstructed needles may further perturb local dose distribution, especially in regions with steep dose gradients. However, as mentioned previously, the automatically detected needles will be visually reviewed and such errors are typically visually apparent and correctable.

\section{Discussion and Conclusion}
In this work, we developed a novel method to automatically localize needles in intraoperative CT for pelvic seed implant brachytherapy. In contrast to existing methods, this method adopted a needle tip-handle detection and matching strategy, which was shown to be more effective to address the challenging imaging characteristics. The experimental results showed our method could achieve a 3D detection F1 score of $90.3 \pm 9.6$, 3D tip-handle position MAE of $1.05 \pm 0.36\,\mathrm{mm}$ and $1.73 \pm 0.95\,\mathrm{mm}$, overall better than the existing most promising segmentation-based method. To the best of our knowledge, our proposed method is the first to employ a detection and matching framework for multi-needle localization.

For the compared segmentation-based method, we chose to utilize the nnUNet model instead of other segmentation methods. As a state-of-the-art, self-configuring framework for medical image segmentation, nnUNet provides a robust baseline by automatically optimizing network parameters for the dataset. This ensures a fairer comparison, focused on the merits of our proposed 3D localization and matching strategies, rather than being influenced by potential limitations of less specialized segmentation networks. Furthermore, to avoid uncertainty introduced by needle separation methods, we chose to manually extract individual needle paths. In this case, the results presented by our tests represent the best possible results obtainable from the segmentation, further emphasizing the advantage of our proposed method.

Although 3D detection may seem intuitive for the first step, 2D detection offers a more practical and efficient solution in our context. 3D detection methods typically demand a substantial dataset for effective training, and manually annotating the tip/handle 3D position for studied CTs is less efficient compared to 2D annotation. Acquiring large, annotated 3D CT datasets can be both challenging and expensive. Furthermore, even if 3D detection were employed, matching needle tips and handles would still be a necessary step to establish the complete 3D needle path.

\noindent\textbf{Future Directions and Practical Implications:}
Several directions may further improve the proposed framework and help mitigate residual risks observed in challenging cases. First, although the current greedy matching and merging strategy provides a favorable balance between accuracy and computational efficiency, it does not guarantee global optimality for the constrained unbalanced assignment problem. Future work will therefore investigate more advanced constrained optimization schemes, such as perturbation-based or neighborhood-search strategies, to further improve solution quality in dense multi-needle scenarios while preserving clinically acceptable runtime \cite{gamezalbanNewPolicyScattered2024}. Second, the current framework assumes approximately linear needle trajectories and relies primarily on static image cues. In practice, however, tissue heterogeneity, puncture resistance, and insertion-induced deformation may cause needle deflection and increase localization uncertainty. A promising direction is to incorporate physics-aware or learning-based models of needle--tissue interaction, so that geometric constraints in the matching stage can be adaptively adjusted according to estimated tissue properties and expected deformation patterns \cite{beheraPrognosisTissueStiffness2024,beheraPredictionPuncturingEvents2024}. Third, broader prospective validation across institutions, scanners, and anatomical sites will be important to further assess robustness, calibrate task-specific priors, and reduce the risk of performance degradation in out-of-distribution clinical settings.

From a practical perspective, the main contribution of this study is to transform needle localization from a fully manual, slice-by-slice process into a rapid computer-assisted workflow. In routine pelvic SIBT, this step is labor-intensive and time-sensitive, especially in cases with a large number of implanted needles. In contrast, the proposed pipeline processes one case in approximately 10 seconds on a standard workstation equipped with a consumer-grade GPU. This modest hardware footprint and near-real-time performance ensure that the framework can be feasibly deployed without prohibitive costs or latency bottlenecks. Following this rapid processing, the reconstructed 3D needle paths can be directly reviewed and, if necessary, minimally adjusted by the physician before dose optimization. This design reduces manual workload, shortens the localization stage, and preserves clinical oversight by positioning the physician as a supervisor rather than a manual delineator. Beyond pelvic SIBT, the proposed tip--handle detection and constrained matching formulation is also potentially transferable to other CT-guided interventions involving straight metallic applicators, such as thoracic or cranial seed implantation\cite{jiangSideEffectsCTguided2018,jiEffectivenessPrognosticFactors2019,LI2024100844}, irreversible electroporation\cite{lakshminarasimhanGlobalSensitivityStudy2023}, and microwave ablation\cite{liMultistageAutomaticRapid2025}, following minor task-specific tuning.

\bibliographystyle{unsrt}
\bibliography{reference}

@inproceedings{zhaoDETRsBeatYOLOs2024a,
  title = {{{DETRs}} beat {{YOLOs}} on real-time object detection},
  booktitle = {2024 {{IEEE}}/{{CVF Conference}} on {{Computer Vision}} and {{Pattern Recognition}} ({{CVPR}})},
  author = {Zhao, Yian and Lv, Wenyu and Xu, Shangliang and Wei, Jinman and Wang, Guanzhong and Dang, Qingqing and Liu, Yi and Chen, Jie},
  year = 2024,
  month = jun,
  pages = {16965--16974},
  publisher = {IEEE},
  address = {Seattle, WA, USA},
  doi = {10.1109/CVPR52733.2024.01605}
}

@inproceedings{huang2025deim,
  title = {{{DeiM}}: {{DETR}} with improved matching for fast convergence},
  booktitle = {Proceedings of the Computer Vision and Pattern Recognition Conference},
  author = {Huang, Shihua and Lu, Zhichao and Cun, Xiaodong and Yu, Yongjun and Zhou, Xiao and Shen, Xi},
  year = 2025,
  pages = {15162--15171}
}

@inproceedings{caiPolyKernelInception2024a,
  title = {Poly kernel inception network for remote sensing detection},
  booktitle = {2024 {{IEEE}}/{{CVF Conference}} on {{Computer Vision}} and {{Pattern Recognition}} ({{CVPR}})},
  author = {Cai, Xinhao and Lai, Qiuxia and Wang, Yuwei and Wang, Wenguan and Sun, Zeren and Yao, Yazhou},
  year = 2024,
  month = jun,
  pages = {27706--27716},
  publisher = {IEEE},
  address = {Seattle, WA, USA},
  doi = {10.1109/CVPR52733.2024.02617}
}

@article{lakshminarasimhanGlobalSensitivityStudy2023,
  title = {Global sensitivity study for irreversible electroporation: Towards treatment planning under uncertainty},
  author = {Lakshmi Narasimhan, Prashanth and Tokoutsi, Zoi and Baroli, Davide and Baragona, Marco and Veroy, Karen and Maessen, Ralph and Ritter, Andreas},
  year = 2023,
  month = mar,
  journal = {Medical Physics},
  volume = {50},
  number = {3},
  pages = {1290--1304},
  doi = {10.1002/mp.16220}
}

@article{beheraPrognosisTissueStiffness2024,
  title = {Prognosis of tissue stiffness through multilayer perceptron technique with adaptive learning rate in minimal invasive surgical procedures},
  author = {Behera, Bulbul and Orlando, M. Felix and Anand, R. S.},
  year = 2024,
  month = may,
  journal = {IEEE Transactions on Medical Robotics and Bionics},
  volume = {6},
  number = {2},
  pages = {769--781},
  doi = {10.1109/TMRB.2024.3377371}
}

@article{jiEffectivenessPrognosticFactors2019,
  title = {The effectiveness and prognostic factors of {{CT}}-guided radioactive {{I-125}} seed implantation for the treatment of recurrent head and neck cancer after external beam radiation therapy},
  author = {Ji, Zhe and Jiang, Yuliang and Tian, Suqing and Guo, Fuxin and Peng, Ran and Xu, Fei and Sun, Haitao and Fan, Jinghong and Wang, Junjie},
  year = 2019,
  month = mar,
  journal = {International Journal of Radiation Oncology*Biology*Physics},
  volume = {103},
  number = {3},
  pages = {638--645},
  doi = {10.1016/j.ijrobp.2018.10.034}
}

@article{LI2024100844,
  title = {A review of the efficacy and safety of iodine-125 seed implantation for lung cancer treatment},
  author = {Li, Zhouzhou and Hu, Zhigang and Xiong, Xiaoqi and Song, Xinyu},
  year = 2024,
  journal = {Cancer Treatment and Research Communications},
  volume = {41},
  pages = {100844},
  doi = {10.1016/j.ctarc.2024.100844}
}

@article{liMultistageAutomaticRapid2025,
  title = {Multi-stage automatic and rapid ablation and needle trajectory planning method for {{CT}}-guided percutaneous liver tumor ablation},
  author = {Li, Shengwei and Zhou, Fanyu and Zhang, Yumeng and Xu, Sheng and Wang, Yufeng and Cheng, Lin and Bie, Zhixin and Li, Bin and Li, Xiao-Guang},
  year = 2025,
  month = jan,
  journal = {Medical Physics},
  volume = {52},
  number = {1},
  pages = {113--130},
  doi = {10.1002/mp.17450}
}

@article{swamidasImageRegistrationContour2020,
  title = {Image registration, contour propagation and dose accumulation of external beam and brachytherapy in gynecological radiotherapy},
  author = {Swamidas, Jamema and Kirisits, Christian and De Brabandere, Marisol and Hellebust, Taran Paulsen and Siebert, Frank-Andr{\'e} and Tanderup, Kari},
  year = 2020,
  month = feb,
  journal = {Radiotherapy and Oncology},
  volume = {143},
  pages = {1--11},
  doi = {10.1016/j.radonc.2019.08.023}
}

@article{czajkowskiAccuracyRegistrationsConebeam2020,
  title = {Accuracy of registrations between cone-beam computed tomography and conventional computed tomography images and dose mapping methods in {{RaySearch}} software for the bladder during brachytherapy of cervical cancer patients},
  author = {Czajkowski, Pawe{\l} and Zwierzchowski, Grzegorz and Piotrowski, Tomasz},
  year = 2020,
  journal = {Journal of Contemporary Brachytherapy},
  volume = {12},
  number = {6},
  pages = {593--600},
  doi = {10.5114/jcb.2020.101693}
}

@article{rodgers3DTransrectalUltrasound2017,
  title = {Toward a {{3D}} transrectal ultrasound system for verification of needle placement during high-dose-rate interstitial gynecologic brachytherapy},
  author = {Rodgers, Jessica Robin and Surry, Kathleen and Leung, Eric and D'Souza, David and Fenster, Aaron},
  year = 2017,
  month = may,
  journal = {Medical Physics},
  volume = {44},
  number = {5},
  pages = {1899--1911},
  doi = {10.1002/mp.12221}
}

@article{weiserAJCC8thEdition2018,
  title = {{{AJCC}} 8th edition: Colorectal cancer},
  author = {Weiser, Martin R.},
  year = 2018,
  month = jun,
  journal = {Annals of Surgical Oncology},
  volume = {25},
  number = {6},
  pages = {1454--1455},
  doi = {10.1245/s10434-018-6462-1}
}

@article{beheraPredictionPuncturingEvents2024,
  title = {Prediction of puncturing events through {{LSTM}} for multilayer tissue},
  author = {Behera, Bulbul and Orlando, M Felix and Anand, R S},
  year = {2024},
  month = oct,
  journal = {Biomedical Physics \& Engineering Express},
  volume = {10},
  number = {6},
  pages = {065041},
  doi = {10.1088/2057-1976/ad844c}
}

@article{gjestebyMetalArtifactReduction2016,
  title = {Metal artifact reduction in {{CT}}: Where are we after four decades?},
  author = {Gjesteby, Lars and De Man, Bruno and Jin, Yannan and Paganetti, Harald and Verburg, Joost and Giantsoudi, Drosoula and Wang, Ge},
  year = 2016,
  journal = {IEEE Access},
  volume = {4},
  pages = {5826--5849},
  doi = {10.1109/ACCESS.2016.2608621}
}

@article{wangInDuDoNetDeepUnfolding2023,
  title = {{{InDuDoNet}}+: A deep unfolding dual domain network for metal artifact reduction in {{CT}} images},
  author = {Wang, Hong and Li, Yuexiang and Zhang, Haimiao and Meng, Deyu and Zheng, Yefeng},
  year = 2023,
  month = apr,
  journal = {Medical Image Analysis},
  volume = {85},
  pages = {102729},
  doi = {10.1016/j.media.2022.102729}
}

@article{karageorgosDenoisingDiffusionProbabilistic2024,
  title = {A denoising diffusion probabilistic model for metal artifact reduction in {{CT}}},
  author = {Karageorgos, Grigorios M. and Zhang, Jiayong and Peters, Nils and Xia, Wenjun and Niu, Chuang and Paganetti, Harald and Wang, Ge and De Man, Bruno},
  year = 2024,
  month = oct,
  journal = {IEEE Transactions on Medical Imaging},
  volume = {43},
  number = {10},
  pages = {3521--3532},
  doi = {10.1109/TMI.2024.3416398}
}

@article{aleongRapidMulticatheterSegmentation2024,
  title = {Rapid Multi-catheter Segmentation for Magnetic Resonance Image-guided Catheter-based Interventions},
  author = {Aleong, Amanda M. and Berlin, Alejandro and Borg, Jette and Helou, Joelle and Beiki-Ardakani, Akbar and Rink, Alexandra and Raman, Srinivas and Chung, Peter and Weersink, Robert A.},
  year = {2024},
  month = aug,
  journal = {Medical Physics},
  volume = {51},
  number = {8},
  pages = {5361--5373},
  issn = {0094-2405, 2473-4209},
  doi = {10.1002/mp.17117},
  urldate = {2025-03-19},
  langid = {english}
}

@article{tongCTguided125IInterstitialBrachytherapy2017,
  title = {{{CT-guided}}{\textsuperscript{125}}{{I}} Interstitial Brachytherapy for Pelvic Recurrent Cervical Carcinoma after Radiotherapy},
  author = {Tong, Lina and Liu, Ping and Huo, Bin and Guo, Zhi and Ni, Hong},
  year = 2017,
  month = aug,
  journal = {OncoTargets and Therapy},
  volume = {10},
  pages = {4081--4088},
  issn = {1178-6930},
  doi = {10.2147/OTT.S139571},
  urldate = {2025-12-06},
  copyright = {http://creativecommons.org/licenses/by-nc/3.0/},
  langid = {english}
}

@article{stoneProstateGlandMotion2002,
  title = {Prostate Gland Motion and Deformation Caused by Needle Placement during Brachytherapy},
  author = {Stone, Nelson N and Roy, Jiten and Hong, Suzanne and Lo, Yeh-Chi and Stock, Richard G},
  year = 2002,
  journal = {Brachytherapy},
  volume = {1},
  number = {3},
  pages = {154--160},
  issn = {15384721},
  doi = {10.1016/S1538-4721(02)00058-2},
  urldate = {2025-12-06},
  langid = {english}
}

@article{DEUFEL2025772,
  title = {How Accurate Is Applicator Reconstruction in {{HDR}} Gynecological Brachytherapy? {{Patient-specific}} Results from an Electromagnetic Tracking System Designed to Intercept Errors before Radiation Delivery},
  author = {Deufel, Christopher L. and Dupere, Justine M. and Brost, Eric E. and Haddock, Michael G. and Garda, Allison E.},
  year = 2025,
  journal = {Brachytherapy},
  volume = {24},
  number = {5},
  pages = {772--782},
  issn = {1538-4721},
  doi = {10.1016/j.brachy.2025.04.007}
}

@article{MILES2008206,
  title = {Equivalent Uniform Dose, {{D90}}, and {{V100}} Correlation with Biochemical Control after Low-Dose-Rate Prostate Brachytherapy for Clinically Low-Risk Prostate Cancer},
  author = {Miles, Edward F. and Nelson, John W. and Alkaissi, Ali K. and Das, Shiva and Clough, Robert W. and Anscher, Mitchell S. and Oleson, James R.},
  year = 2008,
  journal = {Brachytherapy},
  volume = {7},
  number = {2},
  pages = {206--211},
  issn = {1538-4721},
  doi = {10.1016/j.brachy.2008.01.002}
}

@article{jainIntraoperative3DGuidance2012,
  title = {Intra-Operative {{3D}} Guidance and Edema Detection in Prostate Brachytherapy Using a Non-Isocentric {{C-arm}}},
  author = {Jain, A. and Deguet, A. and Iordachita, I. and Chintalapani, G. and Vikal, S. and Blevins, J. and Le, Y. and Armour, E. and Burdette, C. and Song, D. and Fichtinger, G.},
  year = 2012,
  month = apr,
  journal = {Medical Image Analysis},
  volume = {16},
  number = {3},
  pages = {731--743},
  issn = {13618415},
  doi = {10.1016/j.media.2010.07.011},
  urldate = {2025-12-06},
  copyright = {https://www.elsevier.com/tdm/userlicense/1.0/},
  langid = {english}
}

@article{andersenDeepLearningbasedDigitization2020,
  title = {Deep Learning-based Digitization of Prostate Brachytherapy Needles in Ultrasound Images},
  author = {Anders{\'e}n, Christoffer and Ryd{\'e}n, Tobias and Thunberg, Per and Lagerl{\"o}f, Jakob H.},
  year = {2020},
  month = dec,
  journal = {Medical Physics},
  volume = {47},
  number = {12},
  pages = {6414--6420},
  issn = {0094-2405, 2473-4209},
  doi = {10.1002/mp.14508},
  urldate = {2025-02-24},
  langid = {english}
}

@article{barrettArtifactsCTRecognition2004,
  title = {Artifacts in {{CT}}: {{Recognition}} and {{Avoidance}}},
  shorttitle = {Artifacts in {{CT}}},
  author = {Barrett, Julia F. and Keat, Nicholas},
  year = {2004},
  month = nov,
  journal = {RadioGraphics},
  volume = {24},
  number = {6},
  pages = {1679--1691},
  issn = {0271-5333, 1527-1323},
  doi = {10.1148/rg.246045065},
  urldate = {2025-03-19},
  langid = {english}
}

@article{barvaParallelIntegralProjection2008,
  title = {Parallel Integral Projection Transform for Straight Electrode Localization in 3-{{D}} Ultrasound Images},
  author = {Barva, M. and Uhercik, M. and Mari, J.-M. and Kybic, J. and Duhamel, J.-R. and Liebgott, H. and Hlavac, V. and Cachard, C.},
  year = {2008},
  month = jul,
  journal = {IEEE Transactions on Ultrasonics, Ferroelectrics and Frequency Control},
  volume = {55},
  number = {7},
  pages = {1559--1569},
  issn = {0885-3010},
  doi = {10.1109/TUFFC.2008.833},
  urldate = {2025-02-24},
  copyright = {https://ieeexplore.ieee.org/Xplorehelp/downloads/license-information/IEEE.html},
  langid = {english}
}

@article{brayGlobalCancerStatistics2024,
  title = {Global Cancer Statistics 2022: {{GLOBOCAN}} Estimates of Incidence and Mortality Worldwide for 36 Cancers in 185 Countries},
  shorttitle = {Global Cancer Statistics 2022},
  author = {Bray, Freddie and Laversanne, Mathieu and Sung, Hyuna and Ferlay, Jacques and Siegel, Rebecca L. and Soerjomataram, Isabelle and Jemal, Ahmedin},
  year = {2024},
  month = may,
  journal = {CA: A Cancer Journal for Clinicians},
  volume = {74},
  number = {3},
  pages = {229--263},
  issn = {0007-9235, 1542-4863},
  doi = {10.3322/caac.21834},
  urldate = {2025-03-19},
  langid = {english}
}

@inproceedings{carionEndtoEndObjectDetection2020,
  title={End-to-end object detection with transformers},
  author={Carion, Nicolas and Massa, Francisco and Synnaeve, Gabriel and Usunier, Nicolas and Kirillov, Alexander and Zagoruyko, Sergey},
  booktitle={European conference on computer vision},
  pages={213--229},
  year={2020},
  organization={Springer}
}

@article{daiAutomaticMulticatheterDetection2020,
  title = {Automatic Multi-catheter Detection Using Deeply Supervised Convolutional Neural Network in {{MRI}}-guided {{HDR}} Prostate Brachytherapy},
  author = {Dai, Xianjin and Lei, Yang and Zhang, Yupei and Qiu, Richard L.J. and Wang, Tonghe and Dresser, Sean A. and Curran, Walter J. and Patel, Pretesh and Liu, Tian and Yang, Xiaofeng},
  year = {2020},
  month = sep,
  journal = {Medical Physics},
  volume = {47},
  number = {9},
  pages = {4115--4124},
  issn = {0094-2405, 2473-4209},
  doi = {10.1002/mp.14307},
  urldate = {2025-02-24},
  langid = {english}
}

@inproceedings{dai2017deformable,
  title={Deformable convolutional networks},
  author={Dai, Jifeng and Qi, Haozhi and Xiong, Yuwen and Li, Yi and Zhang, Guodong and Hu, Han and Wei, Yichen},
  booktitle={Proceedings of the IEEE international conference on computer vision},
  pages={764--773},
  year={2017}
}

@inproceedings{duanCenterNetKeypointTriplets2019,
  title = {{{CenterNet}}: {{Keypoint Triplets}} for {{Object Detection}}},
  shorttitle = {{{CenterNet}}},
  booktitle = {2019 {{IEEE}}/{{CVF International Conference}} on {{Computer Vision}} ({{ICCV}})},
  author = {Duan, Kaiwen and Bai, Song and Xie, Lingxi and Qi, Honggang and Huang, Qingming and Tian, Qi},
  year = {2019},
  month = oct,
  pages = {6568--6577},
  publisher = {IEEE},
  address = {Seoul, Korea (South)},
  doi = {10.1109/ICCV.2019.00667},
  urldate = {2025-03-20},
  copyright = {https://ieeexplore.ieee.org/Xplorehelp/downloads/license-information/IEEE.html},
  isbn = {978-1-7281-4803-8},
  langid = {english}
}

@inproceedings{galadikovaUsingSimulatedAnnealing2024,
  title = {Using Simulated Annealing for Personnel Assignment in Railway Nodes},
  booktitle = {2024 {{IEEE}} 17th {{International Scientific Conference}} on {{Informatics}} ({{Informatics}})},
  author = {Galad{\'i}kov{\'a}, Andrea and {\v C}asnocha, Branislav},
  year = {2024},
  month = nov,
  pages = {69--74},
  publisher = {IEEE},
  address = {Poprad, Slovakia},
  doi = {10.1109/Informatics62280.2024.10900767},
  urldate = {2025-03-20},
  copyright = {https://doi.org/10.15223/policy-029},
  isbn = {979-8-3503-8768-1},
  langid = {english}
}

@article{gamezalbanNewPolicyScattered2024,
  title = {A New Policy for Scattered Storage Assignment to Minimize Picking Travel Distances},
  author = {G{\'a}mez Alb{\'a}n, Harol Mauricio and Cornelissens, Trijntje and S{\"o}rensen, Kenneth},
  year = {2024},
  month = jun,
  journal = {European Journal of Operational Research},
  volume = {315},
  number = {3},
  pages = {1006--1020},
  issn = {03772217},
  doi = {10.1016/j.ejor.2024.01.013},
  urldate = {2025-03-24},
  langid = {english}
}

@inproceedings{hamzehiCombinatorialReinforcementLearning2019,
  title = {Combinatorial {{Reinforcement Learning}} of {{Linear Assignment Problems}}},
  booktitle = {2019 {{IEEE Intelligent Transportation Systems Conference}} ({{ITSC}})},
  author = {Hamzehi, Sascha and Bogenberger, Klaus and Franeck, Philipp and Kaltenhauser, Bernd},
  year = {2019},
  month = oct,
  pages = {3314--3321},
  publisher = {IEEE},
  address = {Auckland, New Zealand},
  doi = {10.1109/ITSC.2019.8916920},
  urldate = {2025-03-20},
  copyright = {https://ieeexplore.ieee.org/Xplorehelp/downloads/license-information/IEEE.html},
  isbn = {978-1-5386-7024-8},
  langid = {english}
}

@article{hattEnhancedNeedleLocalization2015,
  title = {Enhanced Needle Localization in Ultrasound Using Beam Steering and Learning-Based Segmentation},
  author = {Hatt, Charles R. and Ng, Gary and Parthasarathy, Vijay},
  year = {2015},
  month = apr,
  journal = {Computerized Medical Imaging and Graphics},
  volume = {41},
  pages = {46--54},
  issn = {08956111},
  doi = {10.1016/j.compmedimag.2014.06.016},
  urldate = {2025-03-19},
  langid = {english}
}

@article{hrinivichSimultaneousAutomaticSegmentation2017,
  title = {Simultaneous Automatic Segmentation of Multiple Needles Using {{3D}} Ultrasound for High-Dose-Rate Prostate Brachytherapy},
  author = {Hrinivich, William Thomas and Hoover, Douglas A. and Surry, Kathleen and Edirisinghe, Chandima and Montreuil, Jacques and D'Souza, David and Fenster, Aaron and Wong, Eugene},
  year = {2017},
  month = apr,
  journal = {Medical Physics},
  volume = {44},
  number = {4},
  pages = {1234--1245},
  issn = {00942405},
  doi = {10.1002/mp.12148},
  urldate = {2025-02-24},
  copyright = {http://doi.wiley.com/10.1002/tdm\_license\_1},
  langid = {english}
}

@article{isenseeNnUNetSelfconfiguringMethod2021,
  title = {{{nnU-Net}}: A Self-Configuring Method for Deep Learning-Based Biomedical Image Segmentation},
  shorttitle = {{{nnU-Net}}},
  author = {Isensee, Fabian and Jaeger, Paul F. and Kohl, Simon A. A. and Petersen, Jens and {Maier-Hein}, Klaus H.},
  year = {2021},
  month = feb,
  journal = {Nature Methods},
  volume = {18},
  number = {2},
  pages = {203--211},
  issn = {1548-7091, 1548-7105},
  doi = {10.1038/s41592-020-01008-z},
  urldate = {2025-03-19},
  langid = {english}
}

@article{jiangSideEffectsCTguided2018,
  title = {Side Effects of {{CT-guided}} Implantation of {{125I}} Seeds for Recurrent Malignant Tumors of the Head and Neck Assisted by {{3D}} Printing Non Co-Planar Template},
  author = {Jiang, Yuliang and Ji, Zhe and Guo, Fuxin and Peng, Ran and Sun, Haitao and Fan, Jinghong and Wei, Shuhua and Li, Weiyan and Liu, Kai and Lei, Jinghua and Wang, Junjie},
  year = {2018},
  month = dec,
  journal = {Radiation Oncology},
  volume = {13},
  number = {1},
  pages = {18},
  issn = {1748-717X},
  doi = {10.1186/s13014-018-0959-4},
  urldate = {2025-03-19},
  langid = {english}
}

@article{jungDeeplearningAssistedAutomatic2019,
  title = {Deep-Learning Assisted Automatic Digitization of Interstitial Needles in {{3D CT}} Image Based High Dose-Rate Brachytherapy of Gynecological Cancer},
  author = {Jung, Hyunuk and Shen, Chenyang and Gonzalez, Yesenia and Albuquerque, Kevin and Jia, Xun},
  year = {2019},
  month = oct,
  journal = {Physics in Medicine \& Biology},
  volume = {64},
  number = {21},
  pages = {215003},
  issn = {1361-6560},
  doi = {10.1088/1361-6560/ab3fcb},
  urldate = {2025-02-24},
  langid = {english}
}

@article{kinastHybridMetaheuristicSolution2022,
  title = {A Hybrid Metaheuristic Solution Approach for the Cobot Assignment and Job Shop Scheduling Problem},
  author = {Kinast, Alexander and Braune, Roland and Doerner, Karl F. and {Rinderle-Ma}, Stefanie and Weckenborg, Christian},
  year = {2022},
  month = jul,
  journal = {Journal of Industrial Information Integration},
  volume = {28},
  pages = {100350},
  issn = {2452414X},
  doi = {10.1016/j.jii.2022.100350},
  urldate = {2025-03-20},
  langid = {english}
}

@article{klineWeaponTargetAssignmentProblem2019,
  title = {The {{Weapon-Target Assignment Problem}}},
  author = {Kline, Alexander and Ahner, Darryl and Hill, Raymond},
  year = {2019},
  month = may,
  journal = {Computers \& Operations Research},
  volume = {105},
  pages = {226--236},
  issn = {03050548},
  doi = {10.1016/j.cor.2018.10.015},
  urldate = {2025-03-20},
  langid = {english}
}

@article{kuhnHungarianMethodAssignment1955,
  title = {The {{Hungarian}} Method for the Assignment Problem},
  author = {Kuhn, H. W.},
  year = {1955},
  month = mar,
  journal = {Naval Research Logistics Quarterly},
  volume = {2},
  number = {1-2},
  pages = {83--97},
  issn = {0028-1441, 1931-9193},
  doi = {10.1002/nav.3800020109},
  urldate = {2025-03-19},
  copyright = {http://onlinelibrary.wiley.com/termsAndConditions\#vor},
  langid = {english}
}

@article{wangEfficacyDosimetryAnalysis2020,
  title = {The Efficacy and Dosimetry Analysis of {{CT-guided 125I}} Seed Implantation Assisted with {{3D-printing}} Non-Co-Planar Template in Locally Recurrent Rectal Cancer},
  author = {Wang, Lu and Wang, Hao and Jiang, Yuliang and Ji, Zhe and Guo, Fuxin and Jiang, Ping and Li, Xuemin and Chen, Yi and Sun, Haitao and Fan, Jinghong and Li, Weiyan and Li, Xu and Wang, Junjie},
  year = 2020,
  month = dec,
  journal = {Radiation Oncology},
  volume = {15},
  number = {1},
  pages = {179},
  issn = {1748-717X},
  doi = {10.1186/s13014-020-01607-2},
  urldate = {2025-12-06},
  langid = {english}
}

@inproceedings{law2018cornernet,
  title={Cornernet: Detecting objects as paired keypoints},
  author={Law, Hei and Deng, Jia},
  booktitle={Proceedings of the European conference on computer vision (ECCV)},
  pages={734--750},
  year={2018}
}

@article{linFocalLossDense2020,
  title = {Focal {{Loss}} for {{Dense Object Detection}}},
  author = {Lin, Tsung-Yi and Goyal, Priya and Girshick, Ross and He, Kaiming and Dollar, Piotr},
  year = {2020},
  month = feb,
  journal = {IEEE Transactions on Pattern Analysis and Machine Intelligence},
  volume = {42},
  number = {2},
  pages = {318--327},
  issn = {0162-8828, 2160-9292, 1939-3539},
  doi = {10.1109/TPAMI.2018.2858826},
  urldate = {2025-03-19},
  copyright = {https://ieeexplore.ieee.org/Xplorehelp/downloads/license-information/IEEE.html},
  langid = {english}
}

@inproceedings{liuSSDSingleShot2016,
  title={SSD: Single shot multibox detector},
  author={Liu, Wei and Anguelov, Dragomir and Erhan, Dumitru and Szegedy, Christian and Reed, Scott and Fu, Cheng-Yang and Berg, Alexander C},
  booktitle={European conference on computer vision},
  pages={21--37},
  year={2016},
  organization={Springer}
}

@article{lowekampDesignSimpleITK2013,
  title={The design of SimpleITK},
  author={Lowekamp, Bradley C and Chen, David T and Ib{\'a}{\~n}ez, Luis and Blezek, Daniel},
  journal={Frontiers in neuroinformatics},
  volume={7},
  pages={45},
  year={2013},
  publisher={Frontiers Media SA}
}

@book{maniezzoMatheuristicsAlgorithmsImplementations2021,
  title = {Matheuristics: {{Algorithms}} and {{Implementations}}},
  shorttitle = {Matheuristics},
  author = {Maniezzo, Vittorio and Boschetti, Marco Antonio and St{\"u}tzle, Thomas},
  year = {2021},
  series = {{{EURO Advanced Tutorials}} on {{Operational Research}}},
  publisher = {Springer International Publishing},
  address = {Cham},
  doi = {10.1007/978-3-030-70277-9},
  urldate = {2025-03-20},
  copyright = {https://www.springernature.com/gp/researchers/text-and-data-mining},
  isbn = {978-3-030-70276-2 978-3-030-70277-9},
  langid = {english}
}

@article{martinez-monge125iodineBrachytherapyColorectal1998,
  title = {125iodine Brachytherapy for Colorectal Adenocarcinoma Recurrent in the Pelvis and Paraortics},
  author = {{Mart{\'i}nez-Monge}, Rafael and Nag, Subir and Martin, Edward W},
  year = {1998},
  month = oct,
  journal = {International Journal of Radiation Oncology*Biology*Physics},
  volume = {42},
  number = {3},
  pages = {545--550},
  issn = {03603016},
  doi = {10.1016/S0360-3016(98)00269-7},
  urldate = {2025-03-19},
  copyright = {https://www.elsevier.com/tdm/userlicense/1.0/},
  langid = {english}
}

@article{mazzolaGeneralizedAssignmentNonlinear1989,
  title = {Generalized {{Assignment}} with {{Nonlinear Capacity Interaction}}},
  author = {Mazzola, Joseph B.},
  year = {1989},
  month = aug,
  journal = {Management Science},
  volume = {35},
  number = {8},
  pages = {923--941},
  issn = {0025-1909, 1526-5501},
  doi = {10.1287/mnsc.35.8.923},
  urldate = {2025-03-20},
  langid = {english}
}

@article{mehrtashAutomaticNeedleSegmentation2019,
  title = {Automatic {{Needle Segmentation}} and {{Localization}} in {{MRI With}} 3-{{D Convolutional Neural Networks}}: {{Application}} to {{MRI-Targeted Prostate Biopsy}}},
  shorttitle = {Automatic {{Needle Segmentation}} and {{Localization}} in {{MRI With}} 3-{{D Convolutional Neural Networks}}},
  author = {Mehrtash, Alireza and Ghafoorian, Mohsen and Pernelle, Guillaume and Ziaei, Alireza and Heslinga, Friso G. and Tuncali, Kemal and Fedorov, Andriy and Kikinis, Ron and Tempany, Clare M. and Wells, William M. and Abolmaesumi, Purang and Kapur, Tina},
  year = {2019},
  month = apr,
  journal = {IEEE Transactions on Medical Imaging},
  volume = {38},
  number = {4},
  pages = {1026--1036},
  issn = {0278-0062, 1558-254X},
  doi = {10.1109/TMI.2018.2876796},
  urldate = {2025-02-24},
  copyright = {https://ieeexplore.ieee.org/Xplorehelp/downloads/license-information/IEEE.html},
  langid = {english}
}

@article{nagIntraoperativePlanningEvaluation2001,
  title = {Intraoperative Planning and Evaluation of Permanent Prostate Brachytherapy: {{Report}} of the {{American Brachytherapy Society}}},
  shorttitle = {Intraoperative Planning and Evaluation of Permanent Prostate Brachytherapy},
  author = {Nag, Subir and Ciezki, Jay P and Cormack, Robert and Doggett, Stephen and DeWyngaert, Keith and Edmundson, Gregory K and Stock, Richard G and Stone, Nelson N and Yu, Yan and Zelefsky, Michael J},
  year = {2001},
  month = dec,
  journal = {International Journal of Radiation Oncology*Biology*Physics},
  volume = {51},
  number = {5},
  pages = {1422--1430},
  issn = {03603016},
  doi = {10.1016/S0360-3016(01)01616-9},
  urldate = {2025-03-19},
  copyright = {https://www.elsevier.com/tdm/userlicense/1.0/},
  langid = {english}
}

@article{nguyenCircleRepresentationMedical2022,
  title = {Circle {{Representation}} for {{Medical Object Detection}}},
  author = {Nguyen, Ethan H. and Yang, Haichun and Deng, Ruining and Lu, Yuzhe and Zhu, Zheyu and Roland, Joseph T. and Lu, Le and Landman, Bennett A. and Fogo, Agnes B. and Huo, Yuankai},
  year = {2022},
  month = mar,
  journal = {IEEE Transactions on Medical Imaging},
  volume = {41},
  number = {3},
  pages = {746--754},
  issn = {0278-0062, 1558-254X},
  doi = {10.1109/TMI.2021.3122835},
  urldate = {2025-03-19},
  copyright = {https://ieeexplore.ieee.org/Xplorehelp/downloads/license-information/IEEE.html},
  langid = {english}
}

@article{otsuThresholdSelectionMethod1979,
  title = {A {{Threshold Selection Method}} from {{Gray-Level Histograms}}},
  author = {Otsu, Nobuyuki},
  year = {1979},
  month = jan,
  journal = {IEEE Transactions on Systems, Man, and Cybernetics},
  volume = {9},
  number = {1},
  pages = {62--66},
  issn = {0018-9472, 2168-2909},
  doi = {10.1109/TSMC.1979.4310076},
  urldate = {2025-03-19},
  langid = {english}
}

@article{poloReviewIntraoperativeImaging2010a,
  title = {Review of Intraoperative Imaging and Planning Techniques in Permanent Seed Prostate Brachytherapy},
  author = {Polo, Alfredo and Salembier, Carl and Venselaar, Jack and Hoskin, Peter},
  year = {2010},
  month = jan,
  journal = {Radiotherapy and Oncology},
  volume = {94},
  number = {1},
  pages = {12--23},
  issn = {01678140},
  doi = {10.1016/j.radonc.2009.12.012},
  urldate = {2025-03-19},
  copyright = {https://www.elsevier.com/tdm/userlicense/1.0/},
  langid = {english}
}

@inproceedings{redmonYouOnlyLook2016,
  title = {You {{Only Look Once}}: {{Unified}}, {{Real-Time Object Detection}}},
  shorttitle = {You {{Only Look Once}}},
  booktitle = {2016 {{IEEE Conference}} on {{Computer Vision}} and {{Pattern Recognition}} ({{CVPR}})},
  author = {Redmon, Joseph and Divvala, Santosh and Girshick, Ross and Farhadi, Ali},
  year = {2016},
  month = jun,
  pages = {779--788},
  publisher = {IEEE},
  address = {Las Vegas, NV, USA},
  doi = {10.1109/CVPR.2016.91},
  urldate = {2025-03-20},
  isbn = {978-1-4673-8851-1},
  langid = {english}
}

@article{renFasterRCNNRealTime2017,
  title = {Faster {{R-CNN}}: {{Towards Real-Time Object Detection}} with {{Region Proposal Networks}}},
  shorttitle = {Faster {{R-CNN}}},
  author = {Ren, Shaoqing and He, Kaiming and Girshick, Ross and Sun, Jian},
  year = {2017},
  month = jun,
  journal = {IEEE Transactions on Pattern Analysis and Machine Intelligence},
  volume = {39},
  number = {6},
  pages = {1137--1149},
  issn = {0162-8828, 2160-9292},
  doi = {10.1109/TPAMI.2016.2577031},
  urldate = {2025-03-20},
  copyright = {https://ieeexplore.ieee.org/Xplorehelp/downloads/license-information/IEEE.html},
  langid = {english}
}

@inproceedings{ronnebergerUNetConvolutionalNetworks2015,
  title={U-net: Convolutional networks for biomedical image segmentation},
  author={Ronneberger, Olaf and Fischer, Philipp and Brox, Thomas},
  booktitle={International Conference on Medical image computing and computer-assisted intervention},
  pages={234--241},
  year={2015},
  organization={Springer}
}

@inproceedings{sahuSolvingAssignmentProblem,
  title={Solving the Assignment Problem using Genetic Algorithm and Simulated Annealing.},
  author={Sahu, Anshuman and Tapadar, Rudrajit},
  booktitle={IMECS},
  pages={762--765},
  year={2006}
}

@article{schaferCombiningMachineLearning2023,
  title = {Combining Machine Learning and Optimization for the Operational Patient-Bed Assignment Problem},
  author = {Sch{\"a}fer, Fabian and Walther, Manuel and Grimm, Dominik G. and H{\"u}bner, Alexander},
  year = {2023},
  month = dec,
  journal = {Health Care Management Science},
  volume = {26},
  number = {4},
  pages = {785--806},
  issn = {1386-9620, 1572-9389},
  doi = {10.1007/s10729-023-09652-5},
  urldate = {2025-03-20},
  langid = {english}
}

@article{shaaerDeeplearningassistedAlgorithmCatheter2022,
  title = {Deep-learning-assisted Algorithm for Catheter Reconstruction during {{MR}}-only Gynecological Interstitial Brachytherapy},
  author = {Shaaer, Amani and Paudel, Moti and Smith, Mackenzie and Tonolete, Frances and Ravi, Ananth},
  year = {2022},
  month = feb,
  journal = {Journal of Applied Clinical Medical Physics},
  volume = {23},
  number = {2},
  pages = {e13494},
  issn = {1526-9914, 1526-9914},
  doi = {10.1002/acm2.13494},
  urldate = {2025-03-19},
  langid = {english}
}

@inproceedings{sunDeepHighResolutionRepresentation2019,
  title = {Deep {{High-Resolution Representation Learning}} for {{Human Pose Estimation}}},
  booktitle = {2019 {{IEEE}}/{{CVF Conference}} on {{Computer Vision}} and {{Pattern Recognition}} ({{CVPR}})},
  author = {Sun, Ke and Xiao, Bin and Liu, Dong and Wang, Jingdong},
  year = {2019},
  month = jun,
  pages = {5686--5696},
  publisher = {IEEE},
  address = {Long Beach, CA, USA},
  doi = {10.1109/CVPR.2019.00584},
  urldate = {2025-03-20},
  copyright = {https://ieeexplore.ieee.org/Xplorehelp/downloads/license-information/IEEE.html},
  isbn = {978-1-7281-3293-8},
  langid = {english}
}

@article{wangCTguidedRadioactive125I2023,
  title = {{{CT-guided Radioactive 125I Seed Implantation}} for {{Abdominal Incision Metastases}} of {{Colorectal Cancer}}: {{Safety}} and {{Efficacy}} in 17 {{Patients}}},
  shorttitle = {{{CT-guided Radioactive 125I Seed Implantation}} for {{Abdominal Incision Metastases}} of {{Colorectal Cancer}}},
  author = {Wang, Hao and Shi, Hong-Bing and Qiang, Wei-Guang and Wang, Chao and Sun, Bai and Yuan, Ye and Hu, Wen-Wei},
  year = {2023},
  month = mar,
  journal = {Clinical Colorectal Cancer},
  volume = {22},
  number = {1},
  pages = {136--142},
  issn = {15330028},
  doi = {10.1016/j.clcc.2022.10.004},
  urldate = {2025-03-19},
  langid = {english}
}

@incollection{yangArbitraryOrientedObjectDetection2020,
  title = {Arbitrary-{{Oriented Object Detection}} with {{Circular Smooth Label}}},
  booktitle = {Computer {{Vision}} -- {{ECCV}} 2020},
  author = {Yang, Xue and Yan, Junchi},
  editor = {Vedaldi, Andrea and Bischof, Horst and Brox, Thomas and Frahm, Jan-Michael},
  year = {2020},
  volume = {12353},
  pages = {677--694},
  publisher = {Springer International Publishing},
  address = {Cham},
  doi = {10.1007/978-3-030-58598-3_40},
  urldate = {2025-03-20},
  isbn = {978-3-030-58597-6 978-3-030-58598-3},
  langid = {english}
}

@article{zhangAutomaticMultineedleLocalization2020,
  title = {Automatic Multi-Needle Localization in Ultrasound Images Using Large Margin Mask {{RCNN}} for Ultrasound-Guided Prostate Brachytherapy},
  author = {Zhang, Yupei and Tian, Zhen and Lei, Yang and Wang, Tonghe and Patel, Pretesh and Jani, Ashesh B and Curran, Walter J and Liu, Tian and Yang, Xiaofeng},
  year = {2020},
  month = oct,
  journal = {Physics in Medicine \& Biology},
  volume = {65},
  number = {20},
  pages = {205003},
  issn = {1361-6560},
  doi = {10.1088/1361-6560/aba410},
  urldate = {2025-02-24},
  langid = {english}
}

@article{zhangMultiNeedleDetection3D2020,
  title = {Multi-{{Needle Detection}} in {{3D Ultrasound Images Using Unsupervised Order-Graph Regularized Sparse Dictionary Learning}}},
  author = {Zhang, Yupei and He, Xiuxiu and Tian, Zhen and Jeong, Jiwoong Jason and Lei, Yang and Wang, Tonghe and Zeng, Qiulan and Jani, Ashesh B. and Curran, Walter J. and Patel, Pretesh and Liu, Tian and Yang, Xiaofeng},
  year = {2020},
  month = jul,
  journal = {IEEE Transactions on Medical Imaging},
  volume = {39},
  number = {7},
  pages = {2302--2315},
  issn = {0278-0062, 1558-254X},
  doi = {10.1109/TMI.2020.2968770},
  urldate = {2025-02-24},
  copyright = {https://ieeexplore.ieee.org/Xplorehelp/downloads/license-information/IEEE.html},
  langid = {english}
}

@article{zhengAutomaticNeedleDetection2021,
  title = {Automatic Needle Detection Using Improved Random Sample Consensus in {{CT}} Image-guided Lung Interstitial Brachytherapy},
  author = {Zheng, Yongnan and Jiang, Shan and Yang, Zhiyong and Wei, Lin},
  year = {2021},
  month = apr,
  journal = {Journal of Applied Clinical Medical Physics},
  volume = {22},
  number = {4},
  pages = {121--131},
  issn = {1526-9914, 1526-9914},
  doi = {10.1002/acm2.13231},
  urldate = {2025-02-24},
  langid = {english}
}

@inproceedings{zhouAutomaticNeedleSegmentation2008,
  title = {Automatic Needle Segmentation in {{3D}} Ultrasound Images Using {{3D}} Improved {{Hough}} Transform},
  booktitle = {Medical {{Imaging}}},
  author = {Zhou, Hua and Qiu, Wu and Ding, Mingyue and Zhang, Songgen},
  year = {2008},
  month = mar,
  pages = {691821},
  publisher = {SPIE},
  address = {San Diego, CA},
  doi = {10.1117/12.770077},
  urldate = {2025-02-24},
  langid = {english}
}

@article{zhouDeepLearningbasedAutomatic2024,
  title = {Deep Learning-Based Automatic Pipeline for {{3D}} Needle Localization on Intra-Procedural {{3D MRI}}},
  author = {Zhou, Wenqi and Li, Xinzhou and Zabihollahy, Fatemeh and Lu, David S. and Wu, Holden H.},
  year = {2024},
  month = mar,
  journal = {International Journal of Computer Assisted Radiology and Surgery},
  volume = {19},
  number = {11},
  pages = {2227--2237},
  issn = {1861-6429},
  doi = {10.1007/s11548-024-03077-3},
  urldate = {2025-02-24},
  langid = {english}
}

@article{zhou2019objects,
  title={Objects as points},
  author={Zhou, Xingyi and Wang, Dequan and Kr{\"a}henb{\"u}hl, Philipp},
  journal={arXiv preprint arXiv:1904.07850},
  year={2019}
}

@inproceedings{zhuDeepReinforcementLearning2018,
  title = {Deep {{Reinforcement Learning}} for {{Fairness}} in {{Distributed Robotic Multi-type Resource Allocation}}},
  booktitle = {2018 17th {{IEEE International Conference}} on {{Machine Learning}} and {{Applications}} ({{ICMLA}})},
  author = {Zhu, Qinyun and Oh, Jae},
  year = {2018},
  month = dec,
  pages = {460--466},
  publisher = {IEEE},
  address = {Orlando, FL},
  doi = {10.1109/ICMLA.2018.00075},
  urldate = {2025-03-20},
  isbn = {978-1-5386-6805-4},
  langid = {english}
}

@article{ziliaskopoulosLinearProgrammingModel2000,
  title = {A {{Linear Programming Model}} for the {{Single Destination System Optimum Dynamic Traffic Assignment Problem}}},
  author = {Ziliaskopoulos, Athanasios K.},
  year = {2000},
  month = feb,
  journal = {Transportation Science},
  volume = {34},
  number = {1},
  pages = {37--49},
  issn = {0041-1655, 1526-5447},
  doi = {10.1287/trsc.34.1.37.12281},
  urldate = {2025-03-20},
  langid = {english}
}

@article{zouObjectDetection202023,
  title = {Object {{Detection}} in 20 {{Years}}: {{A Survey}}},
  shorttitle = {Object {{Detection}} in 20 {{Years}}},
  author = {Zou, Zhengxia and Chen, Keyan and Shi, Zhenwei and Guo, Yuhong and Ye, Jieping},
  year = {2023},
  month = mar,
  journal = {Proceedings of the IEEE},
  volume = {111},
  number = {3},
  pages = {257--276},
  issn = {0018-9219, 1558-2256},
  doi = {10.1109/JPROC.2023.3238524},
  urldate = {2025-03-03},
  copyright = {https://ieeexplore.ieee.org/Xplorehelp/downloads/license-information/IEEE.html},
  langid = {english}
}
\end{document}